%% file: main.tex
\documentclass[sigconf]{acmart}

\usepackage[section]{placeins}
\usepackage{amsfonts}
\usepackage{bm}
\usepackage{verbatim}
\usepackage{balance}
\usepackage{graphicx}
\usepackage{multirow}
\usepackage{xspace}
\usepackage{threeparttable}
\usepackage{enumitem}
\usepackage{makecell}
\usepackage{bbding}
\usepackage{caption}
\usepackage{subfig}
\usepackage{booktabs}
\usepackage[export]{adjustbox}
\input{math_commands.tex}

\newcommand{\model}{GraphMAE\xspace}

\newcommand{\vpara}[1]{\vspace{0.04in}\noindent\textbf{#1}\xspace}

\newcommand{\todo}[1]{\textbf{\color{red}[(TODO: #1 )]}}

\newcommand{\hide}[1]{} 
\AtBeginDocument{%
  \providecommand\BibTeX{{%
    \normalfont B\kern-0.5em{\scshape i\kern-0.25em b}\kern-0.8em\TeX}}}

\setcopyright{acmcopyright}

\copyrightyear{2022} 
\acmYear{2022} 
\setcopyright{rightsretained} 

\acmConference[KDD '22]{Proceedings of the 28th ACM SIGKDD Conference on Knowledge Discovery and Data Mining}{August 14--18, 2022}{Washington, DC, USA}
\acmBooktitle{Proceedings of the 28th ACM SIGKDD Conference on Knowledge Discovery and Data Mining (KDD '22), August 14--18, 2022, Washington, DC, USA}
\acmISBN{978-1-4503-9385-0/22/08}
\acmDOI{10.1145/3534678.3539321}

\usepackage{etoolbox}
\makeatletter
\patchcmd{\maketitle}{\@copyrightpermission}{
  \begin{minipage}{0.3\columnwidth}
     \href{https://creativecommons.org/licenses/by/4.0/}{\includegraphics[width=0.90\textwidth]{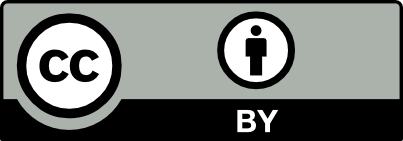}}
  \end{minipage}\hfill
  \begin{minipage}{0.7\columnwidth}
     \href{https://creativecommons.org/licenses/by/4.0/}{This work is licensed under a Creative Commons Attribution International 4.0 License.}
  \end{minipage}
  
  \vspace{5pt}
}{}{}

\makeatother

\begin{document}

\title{\model: Self-Supervised Masked Graph Autoencoders}





\author{Zhenyu Hou}
\affiliation{Tsinghua University}
\email{houzy21@mails.tsinghua.edu.cn}

\author{Xiao Liu}
\affiliation{Tsinghua University}
\email{liuxiao21@mails.tsinghua.edu.cn}

\author{Yukuo Cen}
\affiliation{Tsinghua University}
\email{cyk20@mails.tsinghua.edu.cn}

\author{Yuxiao Dong}
\authornote{Yuxiao Dong and Jie Tang are the corresponding authors.}
\affiliation{Tsinghua University}
\email{yuxiaod@tsinghua.edu.cn}

\author{Hongxia Yang}
\affiliation{DAMO Academy, Alibaba Group}
\email{yang.yhx@alibaba-inc.com}

\author{Chunjie Wang}
\affiliation{BirenTech Research}
\email{cjwang@birentech.com}

\author{Jie Tang}
\authornotemark[1]
\affiliation{Tsinghua University}
\email{jietang@tsinghua.edu.cn}


\renewcommand{\shortauthors}{Zhenyu Hou et al.}

\input{0.abstract}






\maketitle

\renewcommand{\thefootnote}{\fnsymbol{footnote}}
\renewcommand{\thefootnote}{\arabic{footnote}}

\input{1.introduction}
\input{2.related}
\input{3_method_new}
\input{4.experiments}
\input{5.conclusion}

\vpara{Acknowledgements.} This work is supported by Technology and Innovation Major Project of the Ministry of Science and Technology of China under Grant 2020AAA0108400 and 2020AAA0108402, Natural Science
Foundation of China (Key Program, No. 61836013), and National Science
Foundation for Distinguished Young Scholars (No. 61825602).

\balance


\bibliographystyle{ACM-Reference-Format}
\bibliography{reference}

\clearpage
\appendix
\input{appendix}

\end{document}

%% file: math_commands.tex
\usepackage{amsmath,amsfonts,bm}

\def\eqref#1{equation~\ref{#1}}

\def\1{\bm{1}}

\def\vh{{\bm{h}}}

\def\vx{{\bm{x}}}

\def\vz{{\bm{z}}}

\def\mA{{\bm{A}}}

\def\mH{{\bm{H}}}

\def\mX{{\bm{X}}}

\def\mZ{{\bm{Z}}}

\DeclareMathAlphabet{\mathsfit}{\encodingdefault}{\sfdefault}{m}{sl}
\SetMathAlphabet{\mathsfit}{bold}{\encodingdefault}{\sfdefault}{bx}{n}



%% file: 0.abstract.tex
\begin{abstract}

Self-supervised learning (SSL) has been extensively explored in recent years. 
Particularly, generative SSL has seen emerging success in natural language processing and other AI fields, such as the wide adoption of BERT and GPT.
Despite this, contrastive learning---which heavily relies on structural data augmentation and complicated training strategies---has been the dominant approach in graph SSL, 
while the progress of generative SSL on graphs, especially graph autoencoders (GAEs), has thus far not reached the potential as promised in other fields. 
In this paper, we identify and examine the issues that negatively impact the development of GAEs, including their reconstruction objective, training robustness, and error metric. 
We present a masked graph autoencoder \model\footnote{The code is publicly available at \url{https://github.com/THUDM/GraphMAE}.} 
that mitigates these issues for generative self-supervised graph pre-training. 
Instead of reconstructing graph structures, we propose to focus on feature reconstruction with both a masking strategy and scaled cosine error that benefit the robust training of \model. 
We conduct extensive experiments on 21 public datasets for three different graph learning tasks. 
The results manifest that \model---a simple graph autoencoder with careful designs---can 
consistently generate outperformance over both contrastive and generative state-of-the-art baselines. 
This study provides an understanding of graph autoencoders and demonstrates the potential of generative self-supervised pre-training on graphs. 

\end{abstract}

\hide{

Self-supervised learning (SSL) on graph-structured data has been attracting interest.
Generative SSL has been gaining steadily increasing significance due to its success in natural language processing and computer vision.
Despite this, contrastive learning, which relies on complicated training strategies, has been the dominant approach in graph SSL,
while the development of graph generative SSL has far lagged. 
In this paper, we first systematically analyze the issues of existing graph autoencoders and identify the critical components. Based on these, we present a masked graph autoencoder, GraphMAE, for self-supervised graph representation learning. 
The new designs and improvement in \model unleashes the power of autoencoders for graph learning.
We conduct extensive experiments on 21 datasets for three different graph learning tasks: node classification, graph classification, and transfer learning. The results manifest that our proposed method can 
achieve competitive or better performance to state-of-the-art contrastive SSL approaches across three tasks. 
This demonstrates the potential of generative learning in graph.

}

%% file: 1.introduction.tex
\section{Introduction}%
\label{sec:introduction}

\begin{figure*}
    \centering
    \subfloat[\textmd{Technical comparison between generative SSL methods. }]{
        \adjustbox{valign=b}{
        \small
        \renewcommand\tabcolsep{3pt}
        \input{tables/comparsion}
        \label{fig:cmp_tab}}}
    \ 
    \
    \subfloat[\textmd{The effect of \model designs on the performance on Cora dataset. 
    }]{           
        \includegraphics[width=0.48\linewidth,valign=b]{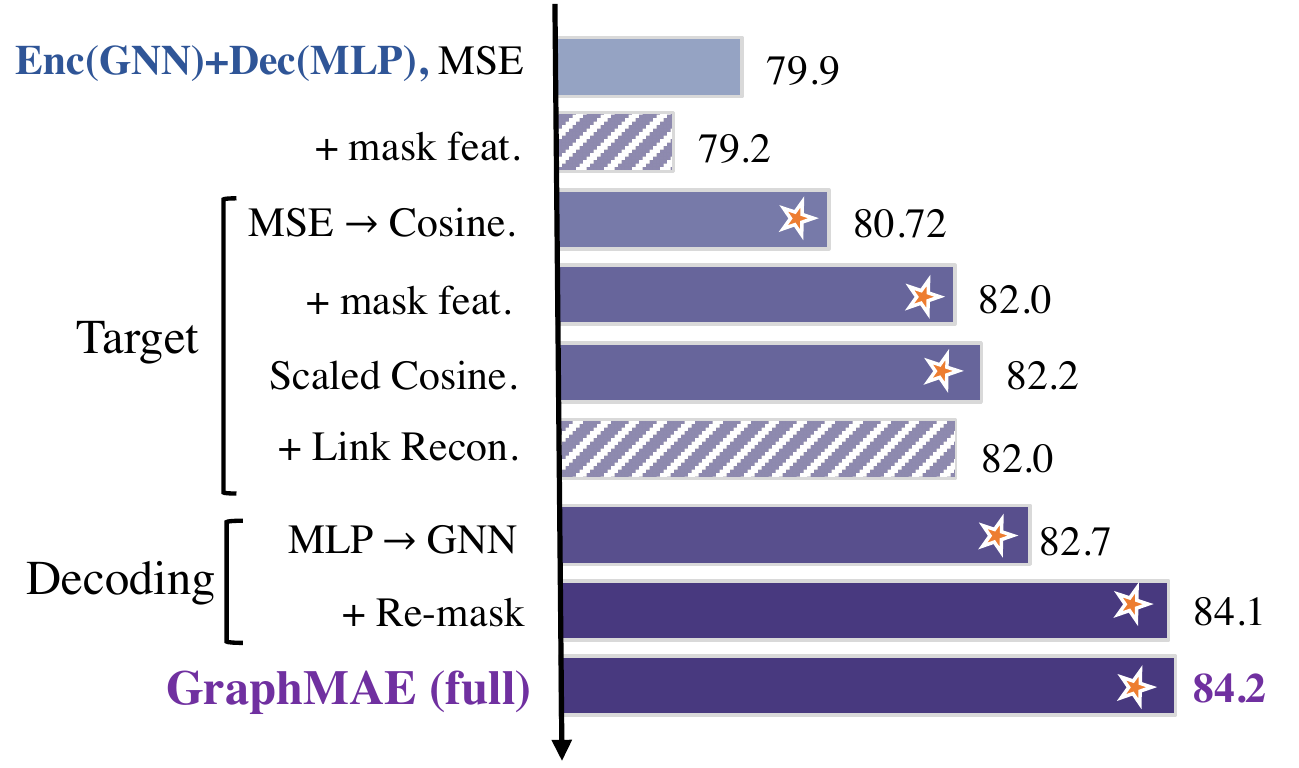} 
        \label{fig:design_eff}          
        }
    \caption{Comparison between generative SSL methods and the effect of \model design. \textmd{
    \textit{AE}: autoencoder methods;
    \textit{No Struct.}: no structure reconstruction objective;
    \textit{Mask Feat.}: use masking to corrupt input features;
    \textit{GNN Decoder}: use GNN as the decoder;
    \textit{Re-mask Dec.}: re-mask encoder output before fed into decoder;
    \textit{Space}:  run-time memory consumption;
\textit{MSE}: Mean Squared Error; 
    \textit{SCE}: Scaled Cosine Error; 
    \textit{CE}: Cross-Entropy Error;
    \textit{SCE} represents our proposed Scaled Cosine Error.}
    }
    \label{fig:cmp_deg}
\end{figure*}

Self-supervised learning (SSL), which can be generally categorized into generative and contrastive methods~\cite{liu2021self}, has found widespread adoption in computer vision (CV) and natural language processing (NLP). 
While contrastive SSL methods have 
 experienced an emergence in the past two years, 
such as MoCo~\cite{he2020momentum}, generative SSL has been gaining steadily increasing significance thanks to several groundbreaking practices, such as the well-established BERT~\cite{devlin2019bert} and GPT~\cite{Radford2019LanguageMA} in NLP as well as the very recent MAE~\cite{he2021masked}  
in CV.

However, in the context of graph learning, 
contrastive SSL has been the dominant approach, especially for two important tasks---node and graph classifications~\cite{velivckovic2018deep, qiu2020gcc,hassani2020contrastive}. 
Its success has been largely built upon relatively complicated training strategies. 
For example, the bi-encoders with momentum updates and exponential moving average are usually indispensable to stabilize the training of GCC~\cite{qiu2020gcc} and BGRL~\cite{thakoor2021bootstrapped}. 
Additionally, negative samples are  necessary for most contrastive objectives, often requiring arduous labor to sample or construct from graphs, e.g., GRACE~\cite{zhu2020deep}, GCA~\cite{zhu2021graph}, and DGI~\cite{velivckovic2018deep}. 
Finally, its heavy reliance on high-quality 
data augmentation proves to be the 
pain point  
of contrastive SSL, e.g., CCA-SSG~\cite{zhang2021canonical}, as graph augmentation is mostly based on heuristics whose effectiveness varies drastically from graph to graph. 


Self-supervised graph autoencoders (GAEs) can naturally avoid the aforementioned issues in contrastive methods, as its learning objective is to directly reconstruct the input graph data~\cite{kipf2016variational,garcia2017learning}.   
Take VGAE for example, it~\cite{kipf2016variational} targets at predicting missing edges. 
EP~\cite{garcia2017learning} instead proposes to recover vertex features. 
GPT-GNN~\cite{hu2020gpt} proposes an autoregressive framework to perform node and edge reconstruction iteratively. 
Later GAEs, 
including ARVGA~\cite{pan2018adversarially}, MGAE~\cite{wang2017mgae}, GALA~\cite{park2019symmetric}, GATE~\cite{amin2020gate}, and AGE~\cite{cui2020adaptive},
majorly focus on the objectives of link prediction and graph clustering. 

\vpara{Dilemmas.} 
Despite their simple forms and various recent attempts 
on them ~\cite{you2020does,jin2020self,hu2020strategies,park2019symmetric,amin2020gate,cui2020adaptive}, the development of self-supervised GAEs has been 
thus far lagged behind contrastive learning. 
To date, there have been no GAEs succeeding to achieve a comprehensive outperformance over contrastive SSL methods, especially on node and graph classifications, which have been significantly advanced by neural encoders, 
e.g., graph neural networks. 
To bridge the gap, we  analyze 
existing GAEs and identify the  issues that may negatively affect the progress of GAEs. 
Note that though previous GAEs may have individually tackled one or two of these issues below, 
none of them collectively deals with the four challenges. 

First, the structure information may be over-emphasized. 
Most  GAEs leverage  link reconstruction as the objective to encourage the topological closeness between neighbors~\cite{kipf2016variational,pan2018adversarially,wang2017mgae,amin2020gate,hu2020gpt,cui2020adaptive}. 
Thus, prior GAEs are usually good at link prediction and node clustering, but unsatisfactory on node and graph classifications. 

Second, feature reconstruction without corruption may not be robust. 
For GAEs~\cite{kipf2016variational,pan2018adversarially,park2019symmetric,amin2020gate,cui2020adaptive} that leverage feature reconstruction, most of them still employ the vanilla architecture that risks learning trivial solutions. 
However, the denoising autoencoders~\cite{vincent2008extracting} that corrupt input and then attempt to recover it have been widely adopted in NLP~\cite{devlin2019bert}, which might be applicable to graphs as well.

Third, 
the mean square error (MSE) can be sensitive and unstable. 
To the best of our knowledge, 
all existing GAEs with feature reconstruction~\cite{wang2017mgae,park2019symmetric,amin2020gate,hu2020gpt,jin2020self} have adopted MSE as the criterion without additional precautions. 
However, MSE is known to suffer from varied feature vector norms and the curse of dimensionality~\cite{Friedman2004OnBV} 
and thus can cause the collapse of training for autoencoders. 

Fourth, 
the decoder architectures are of little expressiveness. 
Most GAEs~\cite{kipf2016variational,pan2018adversarially,wang2017mgae,hu2020strategies,hu2020gpt,cui2020adaptive,jin2020self} leverage  MLP as their decoders. 
However, the targets in language are one-hot vectors containing rich semantics, while in graphs, most of our targets are less informative feature vectors (except for some discrete attributes such as those in chemical graphs). 
In this case, this trivial decoder (MLP) may not be strong enough to bridge the gap between encoder representations and decoder targets for graph features.



\vpara{Contributions.}
In light of the above observations, the goal of this work is to examine to what extent we can 1) mitigate the issues faced by existing GAEs and 2) further enable GAEs to match or outperform contrastive graph learning techniques. 
To this end, we present a masked graph autoencoder \model for self-supervised graph representation learning. 
By identifying the critical components in GAEs, we add new designs and also improve existing strategies for \model, unleashing the power of autoencoders for graph learning. 
Figure \ref{fig:cmp_tab} summarizes the different design choices between GAEs and \model. 
Specifically, the performance of \model largely benefits from the following critical designs  (See Figure \ref{fig:design_eff} for their contributions to performance improvements):

\textbf{Masked feature reconstruction.}
Different from most GAEs' efforts in structure reconstruction, \model only focuses on reconstructing features with masking, whose effectiveness has been extensively verified in CV~\cite{he2021masked} and  NLP~\cite{devlin2019bert,Radford2019LanguageMA}.
Our empirical studies suggest that with a proper error design, masked feature reconstruction can  substantially benefit GAEs.

\textbf{Scaled cosine error.}
Instead of using MSE as existing GAEs, \model employs the cosine error,  which is beneficial when feature vectors vary in their magnitudes (as is often the case for node attributes in graphs). 
On top of it, we further introduce a scaled cosine error to tackle the issue of imbalance between easy and hard samples during reconstruction.

\textbf{Re-mask decoding.}
We leverage a re-mask decoding strategy that re-masks the encoder's output embeddings of masked nodes 
before they are fed into the decoder. 
In addition,  \model proposes to leverage more expressive graph neural nets (GNNs) as its decoder in contrast to previous GAEs' common usage of MLP.


Overall, \model is a simple generative self-supervised  method for graphs without additional cost. 
We conduct extensive experiments on 21 datasets for three different graph learning tasks, including node classification, graph classification, and transfer learning.
The results suggest that equipped with the simple designs above, \model can generate performance advantages over state-of-the-art contrastive SSL approaches across three tasks. 
Moreover, in many cases, \model can match or sometimes outperform supervised baselines, further demonstrating the potential of self-supervised graph learning, particularly generative methods.






\hide{ 

\section{Introduction}%
\label{sec:introduction}
Self-supervised learning (SSL), which can be generally categorized into generative and contrastive methods~\cite{liu2021self}, has found widespread adoption in computer vision (CV), natural language processing (NLP), and graph learning communities. 
While contrastive SSL methods have experienced a sudden emergence in the past two years, 
generative SSL is gaining steadily increasing significance thanks to several successful practices, including well-established BERT~\cite{devlin2019bert}, GPT~\cite{radford2019language} in NLP and very recent MAE~\cite{he2021masked}, iBOT~\cite{zhou2021ibot} in CV.

Generative SSL in graph, including graph autoregressive (GAR) and graph autoencoders (GAEs), has also been studied for long. 
As a typical GAR model, GPT-GNN~\cite{hu2020gpt} proposes an autoregressive framework to perform node and edge reconstruction iteratively. 
For GAEs, VGAE~\cite{kipf2016variational} is a pioneer which targets predicting missing edges.
EP~\cite{garcia2017learning} first propose to recover vertex features. 
Later GAEs, including ARVGA~\cite{pan2018adversarially}, MGAE~\cite{wang2017mgae}, GALA~\cite{park2019symmetric}, GATE~\cite{amin2020gate}, and AGE~\cite{cui2020adaptive} majorly focus on challenges of link prediction and graph clustering. 
Compared to GAR's impractical assumption on graph generating orders, GAEs holds little assumption and thus find more widespread applications.

However, in the context of graph learning, albeit prior efforts on GAEs, 
contrastive SSL is still the dominant approach, especially for node and graph classification. 
Its success has been built upon complex designs and architectures. 
For example, bi-encoders with momentum update, exponential moving average (EMA), and Stop-Gradient are usually indispensable to stabilize the training (e.g., GCC~\cite{qiu2020gcc}, BGRL~\cite{thakoor2021bootstrapped}). 
Negative samples are necessary for most contrastive training objectives, but often require arduous labour to sample or construct from graphs (e.g., GRACE~\cite{zhu2020deep}, GCA~\cite{zhu2021graph}). 
Sometimes, further transformation are needed to create even harder negatives (e.g., DGI~\cite{velivckovic2018deep}). 
Finally, heavy reliance on high-quality and label-invariant data augmentation proves to be the Achilles heel of contrastive SSL (e.g., CCA-SSG~\cite{zhang2021canonical}). 
These augmentation techniques are mostly based on heuristics whose effectiveness vary drastically from graph to graph, and some of them can be unscalable at larger scales (e.g., diffusion in MVGRL~\cite{hassani2020contrastive}).

\vpara{Dilemma.} 
Self-supervised GAEs can naturally avoid aforementioned conflicts of contrastive methods, but suffer from poor performance on node and graph classification. 
Despite a number of recent attempts~\cite{you2020does,manessi2021graph,zhu2020self,jin2020self,hu2020strategies} to handle such challenges with autoencoders, none of them succeed in achieving a comprehensive outperform over state-of-the-art contrastive SSL methods; 
the progress of GAEs in graph has lagged behind NLP and CV. To bridge the gap, we systematically analysis the deficiency of existing GAEs:

(i) Proximity information is over-emphasized. 
Most existing GAEs leverage the structure reconstruction as the main task to encourage topological closeness between neighbour nodes.
Consequently, prior GAEs are usually good at link prediction and clustering, but unsatisfactory on classification.

(ii) Feature reconstruction without corruption is not robust. 
For GAEs that introduce the task of feature reconstruction, 
most of them still employ the vanilla architecture that risks to learn trivial solutions, rather than the more efficient denoising autoencoders~\cite{vincent2008extracting} which corrupt input and then recover.

(iii) Mean Square Error (MSE) criterion can be sensitive and unstable. 
To the best of our knowledge, all the existing GAEs with feature reconstruction have adopted MSE as their criterion. 
However, MSE is known to suffer from varied feature vector norms and curse of dimensionality, and thus can cause the collapse of training. 

(iv) Decoder architectures are of little expressiveness. 
Most GAEs leverage simple MLP as their decoders following BERT. 
However, while in language the targets are one-hot vectors containing rich semantics, in graphs most of our targets are less informative feature vectors (except for some discrete attributes such as in chemical graphs). 
In this case, trivial decoder design may not be strong enough to bridge the gap between encoder representations and decoder targets.

Noted that while previous works on GAEs may have individually tackled one or two of the above defects, none of them jointly deals with the four challenges as a whole to consequently present a unified GAE solution to supersede contrastive ones over extensive benchmarks.

\vpara{Contributions.}
In light of the above observations, we endeavor to revival the idea of generative GAEs by optimizing its design and assimilating the advantages individually appear in prior GAEs. 
We propose \model, a masked graph autoencoder with decoding enhancement, and find that the following crucial designs of \model can lead to substantially boosted performance on node classification, graph classification, and transfer learning: 

\textbf{Masked Feature Reconstruction}: 
\model focuses on feature reconstruction with the more challenging masked prediction objective, 
whose effectiveness and superiority over vanilla architecture has been verified in CV and NLP for long. 

\textbf{Cosine Error}: 
\model employs Cosine Error, which is beneficial when feature vectors vary in their magnitudes (as is often the case for node attributes in graphs), rather than traditional MSE. 
On top of it, we introduce a scaling factor $\gamma\ge 1$ to conquer the problem of imbalance between easy and hard to reconstruct samples.

\textbf{Re-mask Decoding}: 
\model enhance the design of decoder using a more expressive single-layer GNN instead of traditional MLP. 
A novel Re-mask Decoding strategy which re-masks encoder's output embeddings of masked input nodes brings additional improvements.

Surprisingly, without using any complicated tricks including data augmentation, negative sampling, bi-encoders and other techniques to stabilize training, 
\model achieves competitive and even much better performance on a wide range of node classification and graph classification benchmarks compared to state-of-the-art contrastive SSL approaches. 
It also significantly pushes the boundary of graph autoencoders, showing that generative SSL can offer great potential to graph representation learning in the future.

\begin{table}
    \centering
    \caption{Technical comparison between generative SSL methods. \textmd{\textit{AE}: autoencoder methods. \textit{No Struct.}: no structure reconstruction objective. \textit{Mask Feat.}: use masking to corrupt input features. \textit{GNN Decoder}: use GNN as the decoder. \textit{Space} refers to run-time memory consumption. \textit{CE} denotes Cross-Entropy Error.}}
    \vspace{-2mm}
    \small
    \input{tables/comparsion}
    \label{tab:ae_compare}
\end{table}

\begin{figure}
    \centering
    \includegraphics[width=0.48\textwidth]{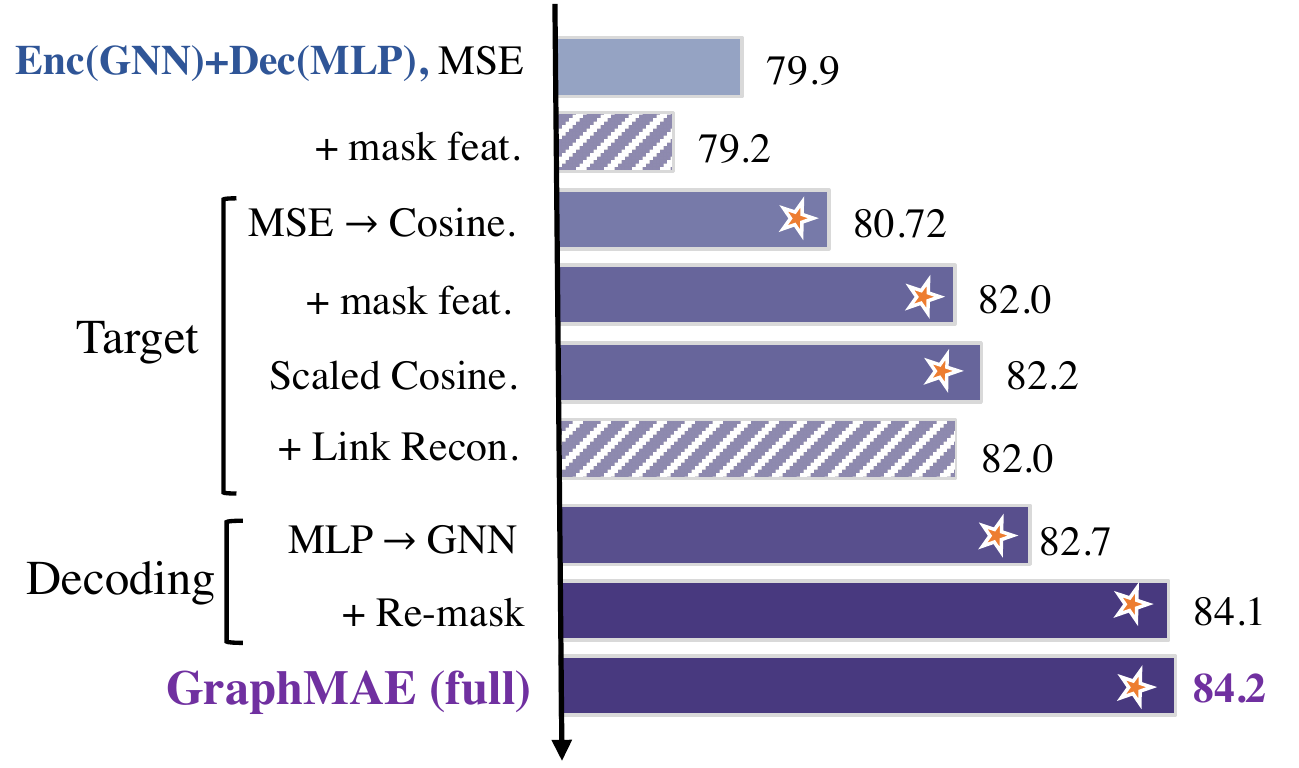}
    \caption{ the effect of \model design. }
    \label{fig:design_xt}
\end{figure}

}

\hide{
\begin{table}[t]
    \centering
    \caption{Technical comparison between \model and contrastive SSL methods. \textmd{Contrastive methods' performance relies on complicated bi-encoder architecture, negative sampling and strong data augmentation; \model do not need them at all.}}
    \vspace{-2mm}
    \small
    \renewcommand\tabcolsep{3pt}
    \input{tables/contrastive_cmp}
    \label{tab:contrastive_compare}
    \vspace{-5mm}
\end{table}
}

\hide{
The idea of masked autoencoders, a form of more general denoising autoencoders, is natural and applicable in computer vision as well. Indeed, closely related research in vision preceded BERT. However, despite signif- icant interest in this idea following the success of BERT, progress of autoencoding methods in vision lags behind NLP. We ask: what makes masked autoencoding different between graph and vision? We attempt to answer this question.

Inspired by the success of contrastive learning in vision, similar methods have been adapted to learning graph neural networks. Although these models have achieved impressive performance, they require complex designs and architectures. For example, DGI and MVGRL rely on a parameterized mutual information estimator to discriminate positive node-graph pairs from negative ones; GRACE and GCA harness an additional MLP-projector to guarantee sufficient capacity. Moreover, negative pairs sampled or constructed from data often play an indispensable role in providing effective contrastive signals and have a large impact on performance. Selecting proper negative samples is often nontrivial for graph-structured data, not to mention the extra storage cost for prohibitively large graphs. BGRL is a recent endeavor on targeting a negative-sample-free approach for GNN learning through asymmetric architectures. However, it requires additional components, e.g., an exponential moving average (EMA) and Stop- Gradient, to empirically avoid degenerated solutions, leading to a more intricate architecture.
}

%% file: tables/comparsion.tex
\begin{tabular}{@{}lccccccc@{}}
\toprule[1.2pt]
Methods  & \makecell[c]{Feat.\\Loss}   & AE         & \makecell[c]{No\\Struc.}  & \makecell[c]{Mask\\Feat.} & \makecell[c]{GNN\\Decoder}    & \makecell[c]{Re-mask\\Dec.}   & Space              \\ \midrule
VGAE~\cite{kipf2016variational} & n/a      & \Checkmark & -          & -          & -             & -    & $\mathcal{O}(N^2)$ \\
ARVGA~\cite{pan2018adversarially}    & n/a      & \Checkmark & -          & -          & -          & -    & $\mathcal{O}(N^2)$ \\
MGAE~\cite{wang2017mgae}     & MSE    & \Checkmark & -          & \Checkmark & -                    & -    & $\mathcal{O}(N)$ \\
GALA~\cite{park2019symmetric}     & MSE    & \Checkmark & \Checkmark & -          & \Checkmark      & -    & $\mathcal{O}(N)$  \\
GATE~\cite{amin2020gate}     & MSE    & \Checkmark & -          & -          & \Checkmark           & -    & $\mathcal{O}(N)$   \\
AttrMask~\cite{hu2020strategies} & CE     & \Checkmark & \Checkmark & \Checkmark & -                & -    & $\mathcal{O}(N)$   \\
GPT-GNN~\cite{hu2020gpt}  & MSE    & -          & -          & \Checkmark & -                       & -    & $\mathcal{O}(N)$   \\
AGE~\cite{cui2020adaptive}     & n/a      & \Checkmark & -          & -          & -                & -    & $\mathcal{O}(N^2)$  \\
NodeProp~\cite{jin2020self} & MSE    & \Checkmark & \Checkmark & \Checkmark & -                     & -    & $\mathcal{O}(N)$ \\
\midrule
\model   & SCE & \Checkmark & \Checkmark & \Checkmark & \Checkmark  & \Checkmark & $\mathcal{O}(N)$ \\ \bottomrule[1.2pt]
\end{tabular}

%% file: tables/contrastive_cmp.tex
\begin{tabular}{@{}clcccc@{}}
\toprule[1.2pt]
SSL                           & Methods & Bi-encoder & NegSample & DataAug    & Space \\ \midrule
\multirow{4}{*}{Contrastive} & MVGRL~\cite{hassani2020contrastive}   & \Checkmark & \Checkmark & \Checkmark & $\mathcal{O}(N^2)$      \\
                              & GCC~\cite{qiu2020gcc}     & \Checkmark & \Checkmark & \Checkmark & $\mathcal{O}(N)$        \\
                              & GraphCL~\cite{you2020graph} & -          & \Checkmark & \Checkmark & $\mathcal{O}(N^2)$      \\
                              & BGRL~\cite{thakoor2021bootstrapped}    & \Checkmark & -          & \Checkmark & $\mathcal{O}(N)$    \\ \midrule
Generative                    & \model                      & -          & -          & -          & $\mathcal{O}(N)$    \\ \bottomrule[1.2pt]
\end{tabular}

%% file: 2.related.tex
\section{Related Work}%
\label{sec:related}
According to model architectures and objective designs, self-supervised methods on graphs can be naturally divided into contrastive and generative domains. 


\subsection{Contrastive Self-Supervised Graph Learning}
Contrastive self-supervised learning
, which encourages alignment between label-invariant distributions and uniformity across other distributions, 
has been the prevalent paradigm for graph representation learning in the last two years. 
Its success relies heavily 
on the elaborate designs of the following components:

\vpara{Negative sampling.}
In pursuit of uniformity, negative sampling is a must for most contrastive methods. 
Mutual information based DGI~\cite{velivckovic2018deep} and InfoGraph~\cite{sun2019infograph} leverage corruptions to construct negative pairs.
GCC~\cite{qiu2020gcc} follows the MoCo-style~\cite{he2020momentum} negative queues. 
GRACE~\cite{zhu2020deep}, GCA~\cite{zhu2021graph} and GraphCL~\cite{you2020graph} use in-batch negatives. 
Despite recent attempts for negative-sample-free contrastive learning, strong regularization from architecture designs (e.g., BGRL ~\cite{thakoor2021bootstrapped}) or 
in-batch feature decorrelation
(e.g. CCA-SSG~\cite{zhang2021canonical}) is necessary in their practice.

\vpara{Architectures.}
Contrastive methods can be unstable in early-stage training, and thus architecture constraints are important to tackle the challenge. 
Asymmetric bi-encoder designs including momentum update~\cite{qiu2020gcc}, EMA, and Stop-gradient~\cite{thakoor2021bootstrapped} are widely adopted.

\vpara{Data augmentation.}
High-quality and informative data augmentation plays a central role in the success of contrastive learning, 
including feature-oriented (partial masking~\cite{jin2020self,hu2020strategies,you2020graph,zhu2020deep,thakoor2021bootstrapped}, shuffling~\cite{velivckovic2018deep}), proximity-oriented (diffusion~\cite{hassani2020contrastive,kefato2021self}, perturbation~\cite{you2020graph,hu2020gpt,zeng2020contrastive}), and graph-sampling-based (random-walk~\cite{qiu2020gcc,hassani2020contrastive,you2020graph}, uniform~\cite{zeng2020contrastive}, ego-network~\cite{sterling2015zinc}) augmentations. 

However, while augmentation in CV is usually human comprehensible, it is difficult to interpret in graphs. 
Without a theoretical understanding of handcrafted graph augmentation strategies, it remains unverified whether they are label-invariant and optimal. 

\subsection{Generative Self-Supervised Graph Learning}
Generative self-supervised learning aims to recover missing parts of the input data. 
It can be further classified into autoregressive and autoencoding two families. 
In previous literature, generative methods' performance on graph representation learning falls behind contrastive methods by a large margin.

\vpara{Graph autoregressive models.}
Autoregressive models decompose joint probability distributions as a product of conditionals. 
In supervised graph generation, previous researchers have proposed GraphRNN~\cite{you2018graphrnn}, GCPN~\cite{you2018graph}. 
For graph representation learning, GPT-GNN~\cite{hu2020gpt} is a recent attempt to leverage graph generation as the training objective. 
However, since most graphs do not present inherent orders, autoregressive methods make little sense on them.

\vpara{Graph autoencoders (GAEs).}
Autoencoders~\cite{hinton1994autoencoders} are designed to reconstruct certain inputs given the contexts and do not enforce any decoding orders as autoregressive methods do. 
The earliest works trace back to GAE and VGAE~\cite{kipf2016variational} which take 2-layer GCN as encoder and dot-product for link prediction decoding. EP~\cite{garcia2017learning} proposes to recover vertex features using mean squared error without input corruption. 
Later GAEs mostly adopt the structural reconstruction (e.g., ARVGA~\cite{pan2018adversarially}) following VGAE, or a combination of structural and feature reconstruction (e.g., MGAE~\cite{wang2017mgae}, GALA~\cite{park2019symmetric} and GATE~\cite{amin2020gate}) as their objectives. And NWR-GAE~\cite{tang2022graph} jointly predicts the node degree and neighbor feature distribution.

Regardless of the successful applications in link prediction and graph clustering, 
due to existing GAEs' reconstruction of structure or/and features without masking, 
their results on node/graph classification benchmarks are usually unsatisfactory.
Therefore, 
our goal in this work is 
to identify the weaknesses of existing GAE designs and rejuvenate the idea of self-supervised GAEs on graph representation learning for classification.

\vpara{\model v.s. Attribute-Masking.}
Recently, some works~\cite{you2020does,manessi2021graph,zhu2020self,jin2020self} have been dedicated to surveying a wide range of many self-supervision objectives' effectiveness on GNNs, including masked feature reconstruction (namely Attribute-Masking).
However, their performance lags far behind state-of-the-art contrastive methods because other critical defects of existing GAEs are not handled. 
We present a rough comparison of algorithms and results between \model and them in Appendix~\ref{app:appendix}.

%% file: 3_method_new.tex
\section{The G\texorpdfstring{\MakeLowercase{raph}}{raph}MAE Approach}

In this section, we present the self-supervised masked graph autoencoder framework---\model---to learn graph representations without supervision based on graph neural networks (GNNs). 
We introduce the critical components that differ \model from previous attempts on designing graph autoencoders (GAEs). 

\begin{figure*}[htbp]
    \centering
    \includegraphics[width=\textwidth]{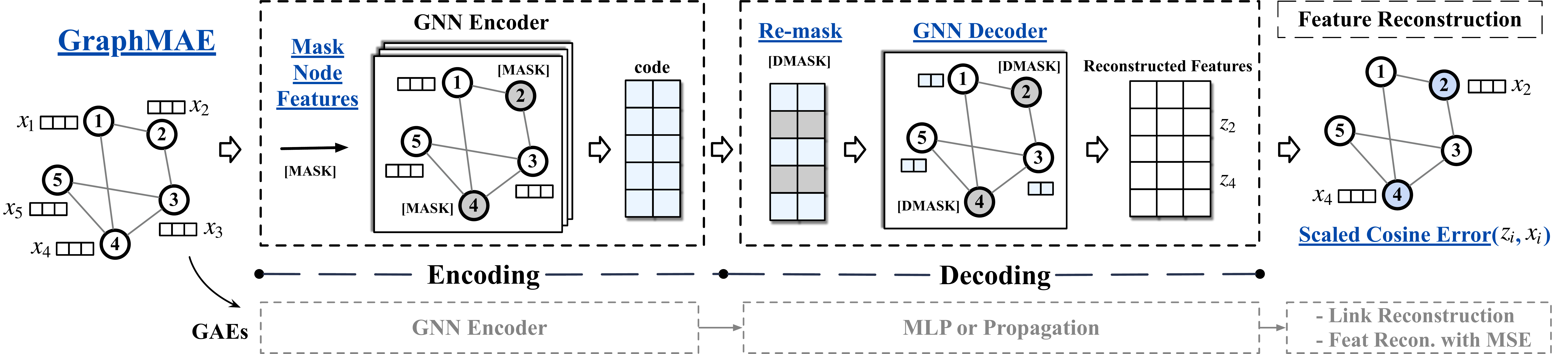}
    \caption{
    Illustration of \model and the comparison with GAE. 
    \textmd{We \underline{underline} the key operations in \model. 
    During pre-training,  \model first masks input node features with a mask token [MASK].  The corrupted graph is encoded into code by a GNN encoder. 
    In the decoding, \model re-masks the code of selected nodes with another token [DMASK], and then employs a GNN, e.g., GAT, GIN, as the decoder. 
    The output of the decoder is used to reconstruct input node features of masked nodes, with the {scaled cosine error} as the criterion.
    Previous GAEs usually use a single-layer MLP or Laplacian matrix in the decoding and focus more on restoring graph structure.}}
    \label{fig:overview}
\end{figure*}

\subsection{The GAE Problem and \model}

Briefly, an autoencoder usually comprises an encoder, code (hidden states), and a decoder. 
The encoder maps the input data to code, and the decoder maps the code to reconstruct the input under the supervision of a reconstruction criterion. 
For graph autoencoders, they can be formalized as follows. 

Let $\mathcal{G}=(\mathcal{V}, \mA, \mX)$ denote a graph, where $\mathcal{V}$ is the node set, $N=|\mathcal{V}|$ is the number of nodes, $\mA \in \{0,1\}^{N\times N}$ is the adjacency matrix, and $\mX \in \mathbb{R}^{N\times d}$ is the input node feature matrix. 
Further, given $f_E$ as the graph encoder, 
$f_D$ as the graph decoder, 
and $\mH \in \mathbb{R}^{N\times d_h}$ denoting the code encoded by the encoder, 
the goal of general GAEs is to reconstruct the input as 
\begin{equation}
    \begin{split}
        \mH = f_E(\mA, \mX), \ \mathcal{G}^{\prime} = f_D(\mA, \mH),
    \end{split}
\end{equation}
where $\mathcal{G}^{\prime}$ denotes the reconstructed graph, which could be either reconstructed features or structures or both.  

Despite their versatile applications in NLP and CV, autoencoders' progress in graphs, especially for classification tasks, is relatively insignificant. 
To bridge the gap, in this work, we aim to identify and rectify the  deficiencies of existing GAE approaches, and subsequently present the \model---a masked graph autoencoder---to further the idea and design of GAEs and generative SSL in graphs.

\vpara{GraphMAE.}
The overall architecture of \model is illustrated in Figure~\ref{fig:overview}. 
Its core idea lies in the reconstruction of masked node features. 
And we introduce a re-mask decoding strategy with GNNs, rather than the widely-used MLP in GAEs, as the decoder to empower \model.
To have a robust reconstruction, we also propose to use a scaled cosine error as the criterion. 
Figure \ref{fig:cmp_tab} summarizes the technical differences between 
\model and existing GAEs.

In detail, the backbones for $f_E$ and $f_D$ can be any type of GNNs, such as GCN~\cite{thomas2017gcn}, GAT~\cite{petar2018gat}, or GIN~\cite{xu2019powerful}. 
As our encoder $f_E$ processes the whole graph $\mA$ with partially observed node features $\widetilde{\mX}$, resonating to the backbones in other generative SSL methods (e.g., BERT and MAE), 
\model prefers a 
more expressive GNN encoder on features for different tasks. 
For instance, GAT is more expressive in node classification, and GIN provides a better inductive bias for graph-level applications (See Tables~\ref{tab:backbone} and ~\ref{tab:ablation}).

\subsection{The   Design of \model}

In this part, we explore how to design a generative self-supervised graph pre-training framework 
that can match and outperform state-of-the-art contrastive models. 
Specifically, we discuss different strategies via answering the following four questions: 
\begin{itemize}[leftmargin=*,itemsep=0pt,parsep=0.2em,topsep=0.3em,partopsep=0.3em]
    \item \textbf{Q1}: What to reconstruct in GAEs? 
    \item \textbf{Q2}: How to train robust GAEs to avoid trivial solutions? 
    \item \textbf{Q3}: How to arrange the decoder for GAEs?
    \item \textbf{Q4}: What error function to use for reconstruction?
\end{itemize}
These questions concern the designs of the reconstruction objective, robust learning, loss function, and model architecture in GAEs, the answers to which enable us to develop \model.

\hide{
\begin{itemize}[leftmargin=*,itemsep=0pt,parsep=0.5em,topsep=0.3em,partopsep=0.3em]
    \item \textbf{Q1}: What should we reconstruct for node/graph classification?
    \item \textbf{Q2}: How to avoid the notorious trivial solution in training GAEs?
    \item \textbf{Q3}: Which criterion to use for feature reconstruction in graphs?
    \item \textbf{Q4}: Why and how do we choose the decoder for GAEs?
\end{itemize}
}

\vpara{Q1: Feature reconstruction as the objective.}
Given a graph $\mathcal{G}=(\mathcal{V}, \mA, \mX)$, a GAE could target reconstructing either  the structure $\mA$ or the features $\mX$, or both of them. 
Most classical GAEs~\cite{kipf2016variational,pan2018adversarially} focus on the tasks of  link prediction and graph clustering, and thus usually choose to reconstruct $\mA$---a target commonly used in network embeddings~\cite{perozzi2014deepwalk,grover2016node2vec,tang2015line}. 
More recent GAEs~\cite{wang2017mgae,park2019symmetric,amin2020gate} tend to adopt a combined objective of reconstructing both features and structure, which unfortunately does not empower GAEs to produce as significant progress in node and graph classifications as autoencoders have done in NLP and CV. 


A very recent study shows that simple MLPs distilled from trained GNN teachers can work comparably to advanced GNNs on node classification~\cite{zhang2022graphless}, indicating the vital role of features in such tasks. 
Thus, to enable \model to achieve a good performance on classification, we adopt feature reconstruction as the  training objective. 
Our empirical examination also shows that the explicit prediction of structural proximity has no contributions to the downstream classification tasks in \model  
(See Figure~\ref{fig:design_eff}).

\vpara{Q2: Masked feature reconstruction.}
When the code's dimension size is larger than input's, the vanilla autoencoders have risks to learn the notorious ``identity function''---the trivial solution---that makes the learned code useless~\cite{vincent2008extracting}. 
Relatively speaking, it is not a severe problem in CV since the image input is usually  high-dimensional. 
However, in graphs, the node feature dimension size is typically quite small, making it a real challenge to train powerful feature-oriented GAEs. 
Unfortunately, existing GAEs that incorporate the reconstruction of features as their objective commonly ignore the threat~\cite{kipf2016variational, pan2018adversarially,park2019symmetric,cui2020adaptive,amin2020gate}. 

The denoising autoencoder~\cite{vincent2008extracting}, which corrupts the input data on purpose, is a natural option to eliminate the trivial solution. 
Actually, the idea of employing masking as the corruption in masked autoencoders has found wide applications in CV~\cite{he2021masked,Bao2021BEiTBP} and NLP~\cite{devlin2019bert}.  
Inspired by their success, we propose to adopt masked autoencoders as the backbone of \model.

Formally, we sample a subset of nodes $\widetilde{\mathcal{V}}\subset\mathcal{V}$ and mask each of their features with a mask token [MASK], i.e., a learnable vector $\vx_{[M]}\in \mathbb{R}^{d}$.
Thus, the node feature $\widetilde{\vx}_{i}$ for $v_i\in\mathcal{V}$ in the masked feature matrix $\widetilde{\mX}$ can be defined as:
$$
\widetilde{\vx}_{i}=
\begin{cases}
\vx_{[M]} & v_i\in\widetilde{\mathcal{V}} \\
\vx_i & v_i\notin\widetilde{\mathcal{V}} 
\end{cases}
$$
The objective of \model is to reconstruct the masked features of nodes in $\widetilde{\mathcal{V}}$ given the partially observed node signals $\widetilde{\mX}$ and the input adjacency matrix $\mA$.

We apply a uniform random sampling strategy without replacement to obtain masked nodes. 
In GNNs, each node relies on its neighbor nodes to enhance/recover its features~\cite{gilmer2017neural}. 
Random sampling with a uniform distribution helps prevent a potential bias center, i.e., one's neighbors are neither all masked nor all visible. 
Additionally, 
similar to MAE~\cite{he2021masked}, 
a relatively large mask ratio (e.g., 50\%) is necessary to reduce redundancy in the attributed graphs in most cases
and thus form a challenging self-supervision to learn meaningful node representations. 

The use of [MASK], on the other hand, potentially creates a mismatch between training and inference since the [MASK] token does not appear during inference~\cite{yang2019xlnet}. 
To mitigate the discrepancy, 
BERT proposes to not always replace ``masked'' words with the actual [MASK] token, but with a small probability (i.e., 15\% or smaller) to leave it unchanged or to substitute it with another random token. 
Our experiments find that the ``leave-unchanged'' strategy actually harms \model's learning, while the ``random-substitution'' method could 
help form more high-quality representations.

\vpara{Q3: GNN decoder with re-mask decoding.}
The decoder $f_D$ maps the latent code $\mH$ back to the input $\mX$, 
and its design 
would
depend on the semantic level~\cite{he2021masked} of target $\mX$. 
For example, in language, since targets are one-hot missing words with rich semantics, usually a trivial decoder such as MLP is sufficient~\cite{devlin2019bert}. 
But in vision,  previous studies~\cite{he2021masked} discover that a more advanced decoder (e.g., the Transformer model~\cite{vaswani2017attention}) is necessary to recover pixel patches with low-level semantics.

In graphs, the decoder reconstructs relatively less informative multi-dimensional node features. 
Traditional GAEs employ either no neural decoders or a simple MLP for decoding with less expressiveness, causing the latent code $\mH$ to be nearly identical to input features. 
However, it has no merit to learn such trivial latent representations because the goal is to embed input features with  meaningful compressed knowledge. 
Therefore, \model resorts to a more expressive single-layer GNN as its decoder. 
The GNN decoder can recover the input features of one node based on a set of nodes instead of only the node itself, and it consequently helps the encoder learn high-level latent code. 

To further encourage the encoder to learn compressed representations, we propose a \textit{re-mask decoding} technique to process the latent code $\mH$ for decoding. 
We replace $\mH$ on masked node indices again with another mask token [DMASK], i.e., the decoder mask, with $\vh_{[M]} \in \mathbb{R}^{d_h}$. 
Specifically, the re-masked code $\widetilde{\vh}_i$ in $\widetilde{\mH} = \mathrm{REMASK}(\mH)$ can be denoted as 
$$
\widetilde{\vh}_i = 
    \begin{cases}
        \vh_{[M]} &  v_i \in \widetilde{\mathcal{V}} \\
        \vh_i  &   v_i \notin \widetilde{\mathcal{V}} 
    \end{cases}
$$
With the GNN decoder, a masked node is forced to reconstruct its input feature from the neighboring unmasked latent representations. 
Similar to encoders, our empirical examination suggests that the GAT and GIN encoders are good options for node classification and  graph classification, respectively. 
Note that the decoder is only used during the self-supervised training stage to perform the node feature reconstruction task. 
Therefore, the decoder architecture is independent of the encoder choice and can use any type of GNN.

\vpara{Q4: Scaled cosine error as the criterion.}
The feature reconstruction criterion varies 
for masked autoencoders~\cite{devlin2019bert,he2021masked} in different domains. 
In NLP and CV, the \textit{de facto} criterion is to predict discrete token indices derived from tokenizers using cross entropy error. 
An exception is the MAE work~\cite{he2021masked} in CV, which directly predicts pixels in the masked patches using the mean square error (MSE); 
Nevertheless in fact, pixels are naturally normalized to 0--255, functioning similarly to tokenizers. 
But in graphs, it remains unexplored how to define a universal tokenizer. 

In \model, we propose to directly reconstruct the raw features for each masked node, which can be challenging due to the multi-dimensional and continuous nature of node features. 
Existing GAEs with feature reconstruction have adopted MSE as their criterion~\cite{wang2017mgae,park2019symmetric,jin2020self}. 
But in preliminary experiments, we discover that 
the MSE loss can be minimized to nearly zero
and may not be enough for feature reconstruction, which may (partly) explain why few existing GAEs use feature reconstruction as their only training objective. 
To this end, we found that MSE could suffer from the issues of \textit{sensitivity} and \textit{low selectivity}. Sensitivity means that MSE is sensitive to vector norms and dimensionality~\cite{Friedman2004OnBV}. Extreme values in certain feature dimensions can also lead to MSE’s overfit on them. Low selectivity represents that MSE is not selective enough to focus on those harder ones among imbalanced easy-and-hard samples. 

To handle its sensitivity, we leverage the cosine error as the criterion to reconstruct original node features, which gets rid of the impact of dimensionality and vector norms. 
The $l_2$-normalization in the cosine error maps vectors to a unit hyper-sphere and substantially improves the training stability of representation learning. 
This benefit is also observed by some contrastive learning methods like BYOL~\cite{grill2020bootstrap}. 

To improve its selectivity, we further the cosine error by introducing the scaled cosine error (SCE) for \model. 
The intuition is that we can down-weight easy samples' contribution in training by scaling the cosine error with a power of $\gamma\ge 1$. 
For predictions with high confidence, their corresponding cosine errors are usually smaller than 1 and decay faster to zero when the scaling factor $\gamma>1$. 
Formally speaking, given the original feature $\mX$ and reconstructed output $\mZ=f_D(\mA, \widetilde{\mH})$, we define SCE for \model as 
\begin{equation}
    \mathcal{L}_{\textrm{SCE}} = \frac{1}{|\widetilde{\mathcal{V}}|} \sum_{v_i \in \widetilde{\mathcal{V}}} (1 - \frac{\vx^T_i \vz_i}{\| \vx_i\| \cdot \| \vz_i\|})^\gamma,~ \gamma \ge 1,
    \label{eq:loss}
\end{equation}
which is averaged over all masked nodes. 
The scaling factor $\gamma$ is a hyper-parameter adjustable over different datasets.
This scaling technique could also be viewed as an adaptive sample reweighing, and the weight of each sample is adjusted with the reconstruction error. 
This error is also famous in the field of supervised object detection as the focal loss~\cite{lin2017focal}.




\hide{Intuitively, using an MLP decoder is more likely to reconstruct a node's input feature from its encoded embedding. As a method of self-supervised learning, feature reconstruction aims to help learn expressive node representations, instead of making the encoded representations similar to raw node features. To achieve this target, we found that the decoder design plays an important role in learning meaningful representations. }

In summary, \model is a simple and scalable self-supervised graph learning framework. 
Figure \ref{fig:cmp_deg} illustrates how each of its design choices directly impacts the performance of the self-supervised \model framework. 
By identifying the negative components and designing new strategies, \model unleashes the power of autoencoders for self-supervised graph pre-training.


\subsection{Training and Inference}
The overall training flow of \model  is summarized by Figure \ref{fig:overview}.
First, given an input graph, we randomly select a certain proportion of nodes and replace their node features with the mask-token [MASK]. 
We feed the graph with partially observed features into the encoder to generate the encoded node representations. 
In decoding, we re-mask the selected nodes and replace their features with another token [DMASK]. Then the decoder is applied to the re-masked graph to reconstruct the original node features with the proposed scaled cosine error.

For downstream applications, the encoder is applied to the input graph without any masking in the inference stage. 
The generated node embeddings can be used for various graph learning tasks, such as node classification and graph classification. 
For graph-level tasks, we use a non-parameterized graph pooling (readout) function, e.g., MaxPooling and  MeanPooling, to obtain the graph-level representation 
$ \vh^g = \mathrm{READOUT}(\{\vh_i, v_i \in \mathcal{G}_g \}) $.
In addition, similar to ~\cite{hu2020strategies}, \model also enables robust transfer of pre-trained GNN models to various downstream tasks.
In the experiments, we show that \model achieves competitive performance in both node-level and graph-level applications.


\hide{

\section{\model: Self-supervised Masked Graph Autoencoder}
\begin{figure*}[htbp]
    \centering
    \includegraphics[width=\textwidth]{imgs/overview_small.pdf}
    \caption{Illustration of \model and comparison with GAE. \model corrupts input graph by masking node features to create a challenging objective. The graph is encoded into latent representation by GNN encoder.  \model would re-mask the encoding of nodes before fed into the decoder, and employ a GNN, e.g. GAT, GIN, as the decoder. 
    The output of decoding is used to reconstruct input node features of masked nodes, with \textit{scaled cosine error} as the criterion.
    Previous GAEs usually use a linear transformation or MLP in the decoding, and more focus on restoring graph structure.}
    \label{fig:overview}
\end{figure*}

The idea of autoencoders, which comprise the encoder, code, and the decoder, has been popular in representation learning for decades. The encoder maps input data to code, and the decoder maps the code back for input reconstruction under the supervison of a reconstruction creterion. To acquire a compressed knowledge representation, autoencoders usually impose a bottleneck in their architecture by enforcing a smaller dimension in code than the input.

Despite versatile applications in NLP and CV, autoencoders' progress in graphs (especially for classification) is slow in recent years. To bridge the gap, in this work we want to identify and rectify crucial deficiencies of existing approaches, and consequently present \model, a masked graph autoencoder framework, to renew the idea and design of GAEs and generative SSL in graphs.

Consider a graph $\mathcal{G}=(\mathcal{V}, \mA, \mX)$, $\mathcal{V}$ is node set, $\mA \in \{0,1\}^{N\times N}$ is the adjacency matrix, and $\mX \in \mathbb{R}^{N\times d}$ is the original node feature matrix. Let $f_E$ refers to the graph encoder, $f_D$ denotes the graph decoder, $\mH \in \mathbb{R}^{N\times d_h}$ denotes to the code (or hidden states) encoded by the encoder, and $\mZ \in \mathbb{R}^{N\times d}$ denotes reconstructed features. Given masked feature matrix $\widetilde{\mX}$, \model reconstructs inputs as
\begin{equation}
    \begin{split}
        \mH = f_E(\mA, \widetilde{\mX}), \ \mZ = f_D(\mA, \mH)
    \end{split}
\end{equation}

\noindent where the backbone for $f_E,f_D$ can be any type of GNN, such as GCN~\cite{thomas2017gcn}, GAT~\cite{petar2018gat}, or GIN~\cite{xu2019powerful}. As our encoder $f_E$ processes the whole graph $\mA$ with partially observed node features $\widetilde{\mX}$, resonating to the backbones in other generative SSL methods (e.g., BERT and MAE), \model may prefer a more expressive encoder on features. For instance, GAT is more expressive in node classification, and GIN provides a better inductive bias for graph level applications. We present extensive studies on backbones for encoder in Table~\ref{tab:backbone} and decoder in Table~\ref{tab:ablation}.

\subsection{Critical Designs of \model}
In this part, we delve into the why and wherefores of \model being more efficient and effective compared to previous GAEs. Particularly, we will introduce our ideas via answering the following four questions

\begin{itemize}[leftmargin=*,itemsep=0pt,parsep=0.5em,topsep=0.3em,partopsep=0.3em]
    \item \textbf{Q1}: What should we reconstruct for node/graph classification?
    \item \textbf{Q2}: How to avoid the notorious trivial solution in training GAEs?
    \item \textbf{Q3}: Which criterion to use for feature reconstruction in graphs?
    \item \textbf{Q4}: Why and how do we choose the decoder for \model?
\end{itemize}

which concerns the designs of objective, learning, loss function, and model architecture.

\vpara{Q1: Feature reconstruction as the objective.}~\\

For a graph $\mathcal{G}=(\mathcal{V}, \mA, \mX)$, $\mathcal{V}$, a GAE could target reconstructing either the $\mA$, $\mX$, or both of them. Classical GAEs usually choose to reconstruct $\mA$ following the fashion in early era of network embedding~\cite{perozzi2014deepwalk,grover2016node2vec,tang2015line}, when research focus was on the topological closeness and attributes seldom appeared in acknowledged benchmarks; thus, no feature was needed to be reconstructed. Besides, designing proper criterion to predict binary linking is much easier than node features, which can be potentially multi-dimensional and continuous.

However, in the era of GNNs, features and their entanglement with structures may count more, especially for node and graph classification challenges. Considering this paradigm shift, we argue that for a better performance on classification, feature reconstruction should be alternatively chosen as the GAE training objectives. Based on this intuition, next we introduce how we prevent trivial solution in the training of feature-oriented GAEs.

\vpara{Q2: More challenging masked feature reconstruction.}~\\
When code's dimension is larger than input's, vanilla autoencoders are risking to learn the notorious ``identity function'' that makes the learned code useless. It is not an extremely severe problem in CV since the image input is high-dimensional; but in graphs, feature dimension is quite low and thus making it a real challenge to feature-oriented GAEs. Unfortunately, existing GAEs that incorporates feature reconstruction as objectives usually ignore the threat.

Denoising autoencoder~\cite{vincent2008extracting}, which corrupts the input data on purpose, is a natural solution to the challenge of trivial solution. Typically, the idea of masked autoencoders to employ masking as the corruption has found wide applications in CV and NLP. Inspired by its success, we propose to adopt masked autoencoders as the backbone for self-supervised representation learning on graphs, namely the \model.

 Formally, we sample a subset of nodes $\widetilde{\mathcal{V}}\subset\mathcal{V}$ and mask the features of nodes in $\widetilde{\mathcal{V}}$ with a learnable mask token [MASK] (i.e., $\vx_m\in \mathbb{R}^{d}$). Thus, we define node feature $\widetilde{\vx}_{i}$ for $v_i\in\mathcal{V}$ in masked feature matrix $\widetilde{\mX}$ as:
\begin{equation}
\widetilde{\vx}_{i}=
\begin{cases}
x_m & v_i\in\widetilde{\mathcal{V}} \\
x_i & v_i\notin\widetilde{\mathcal{V}} 
\end{cases}
\end{equation}
 
\noindent Our objective is to reconstruct the masked node features given the partial observed node signals and adjacency matrix.  

We apply a uniform random sampling without replacement strategy to obtain masked nodes. In GNNs, each node relies on neighboring nodes to enhance/recover its feature. Random sampling follows a uniform distribution and helps prevents a potential bias center (i.e. one's neighbors are neither all masked nor all visible). Additionally, unlike BERT~\cite{devlin2019bert} in NLP but resonating to MAE~\cite{he2021masked} in CV, in most cases a relatively large mask ratio (i.e., 50\%) is necessary to reduce redundancy in attributed graph and create a challenging self-supervision to learn meaningful node representations. 

The use of [MASK], on the other hand, potentially creates a mismatch between training and
inference, since the [MASK] token does not appear during inference~\cite{yang2019xlnet}. To mitigate the discrepancy and strength \model's robustness, BERT propose to not always replace ``masked'' nodes with the actual [MASK] token, but with a small probability (i.e., 15\% or smaller) to leave it unchanged, or substitute it with another random token. In our experiments, we find that ``leaving-unchanged'' strategy harms \model's learning, while the ``random-substitution'' can sometimes help to form a more robust and better-performed representation. 

\vpara{Q3: Scaled Cosine Error as the criterion.}~\\
Reconstruction criterion counts for masked autoencoders. In NLP and CV, the de facto criterion is to predict discrete token indices derived from tokenizers using CrossEntropy Error. An exception would be MAE~\cite{he2021masked} in CV, which directly predict pixels in the masked patches using Mean Square Error (MSE); but it highly depends on the fact that pixels are naturally normalized to 0-255, functioning similarly to tokenizers. But in graph, it remains unexplored how to define a universal tokenizer. 

As a result, we propose to directly reconstruct the raw features for each masked node, which can be challenging owing to the multi-dimensional and continuous nature of node features. 
Existing GAEs with feature reconstruction have all adopted MSE as their criterion. But in preliminary experiments, we discover that without effort to lift a finger, MSE loss can be minimized to nearly zero. Therefore, the MSE is not enough for feature reconstruction, which explains why none of existing GAEs use feature reconstruction as their only training objective. After analysis, we believe that MSE has two severe shortcomings for self-supervised graph learning:

\begin{itemize}[leftmargin=*,itemsep=0pt,parsep=0.5em,topsep=0.3em,partopsep=0.3em]
    \item \textbf{Sensitivity}: MSE, or Euclidean Distance, is sensitive to vector norms and dimensionality. Extreme values in certain feature dimensions can also lead to MSE's overfit on them.
    \item \textbf{Low selectivity}: we find majority of node features are easy to recover because of their high expressiveness, which results in an extreme imbalance between easy and hard samples. MSE is not selective enough to focus \model on hard ones.
\end{itemize}

To handle the sensitivity, we leverage Cosine Error as the criterion to reconstruct original node features, which gets rid of impact from dimensionality and vector norms. 
$l_2$-normalization in cosine error maps vectors to a unit hyper-sphere and substantially improves training stability of representation learning. This benefit is also observed by some contrastive learning methods like BYOL~\cite{grill2020bootstrap}. 

On top of Cosine Error, we propose Scaled Cosine Error (SCE) to improve the selectivity. The intuition is that we can down-weight easy samples' contribution in the training by scaling Cosine Error with a power of $\gamma\ge 1$. For those prediction with high confidence, their corresponding cosine error is usually smaller than 1 and decays faster to zero when scaling factor $\gamma>1$. Formally speaking, given original input feature $\vx$ and reconstructed output $\vz$, we define SCE for \model as
\begin{equation}
    \mathcal{L} = -\frac{1}{|\widetilde{\mathcal{V}}|} \sum_{v_i \in \widetilde{\mathcal{V}}} (1 - \frac{\vx^T_i \vz_i}{\| \vx_i\| \cdot \| \vz_i\|})^\gamma,~ \gamma \ge 1
    \label{eq:loss}
\end{equation}
\noindent which is averaged over all masked nodes. Scaling factor $\gamma$ is a hyper-parameter adjustable over different datasets.
This scaling technique could be viewed as adaptive sample reweighing, and the weight of each sample is adjusted with the reconstruction error. It is also famous in the field of supervised object detection as the focal loss~\cite{lin2017focal}.

\hide{
The gradient of SCE is:
\begin{equation}
\nabla L^\alpha_{\theta, \phi} = \alpha L^{\alpha-1} \nabla L_{\theta, \phi}
\label{eq:grad}
\end{equation}
The above equation indicates that, on the one hand, $L^{\alpha-1}$ scales the gradient based on the the reconstruction error of ($\vx$, $\vz$). A sample would be assigned higher weight relatively if its reconstruction error $L$ is high. On the other hand, if $L < 1 (\vx^\vz > 0)$, that is, the reconstructed $\vz$ tends to be similar to node feature $\vx$, SCE would lower it weight during training. This makes optimization focus more on hard samples and helps convergence. (??)

Eq \ref{eq:loss} looks like the criterion in BYOL~\cite{grill2020bootstrap}. But there exists obvious differences. (i) Learning objective. BYOL contrasts representations of different augmented views, which are from two different networks (online network and target network). \model aims to reconstruct the input from the output of the decoder. (ii)  The targte network and representation is constantly updating as the training proceeds, while the reconstructed target of \model is fixed (the original node feature). }

\vpara{Q4: GNN Decoder with Re-mask Decoding.}~\\
The decoder $f_D$ maps the latent code $\mH$ back to the input $\mX$, and its complexity can highly depends on targets' semantic level. In language, since targets are one-hot missing words with rich semantics, usually a trivial decoder such as MLP is sufficient~\cite{devlin2019bert}. But in vision, when previous works~\cite{he2021masked} discover that a more complicated decoder based on transformers is necessary to recover pixel patches with low-level semantics.

In graph, the decoder reconstructs less informative multi-dimensional node features. 
Traditional GAEs employ no decoder or a simple MLP for decoding with little expressiveness, causing the latent code $\mH$ to be nearly identical to input features. However, it has no merit to learn such trivial representations because what we want are embeddings with meaningful compressed knowledge. Therefore, \model resorts to a more expressive single-layer GNN as its decoder. The GNN decoder recovers input features based on a set of nodes instead a node itself, and consequently helps the encoder learn high-level latent code.

To further encourage encoder learning compressed representation, we propose a novel technique \textit{Re-mask Decoding} to process latent code $\mH$ for decoding. We replace the $\mH$ again on masked node indices with another mask token [DMASK] (i.e., decoder mask) with $\vh_m \in \mathbb{R}^{d_h}$. Denote the re-masked results as $\widetilde{\mH} = \mathrm{REMASK}(\mH)$
$$
\widetilde{\vh}_i = 
    \begin{cases}
        \vh_m, &  v_i \in \widetilde{\mathcal{V}} \\
        \vh_i &   v_i \notin \widetilde{\mathcal{V}} 
    \end{cases}
$$
The [DMASK] is shared over all nodes and different from the mask token used in encoding. 
With the GNN decoder, a masked node is forced to reconstruct its input feature from the neighboring unmasked latent representations. In our experiments, we find that GAT decoder performs better for node classification and GIN decoder is a good choice for graph classification. 

\hide{Intuitively, using an MLP decoder is more likely to reconstruct a node's input feature from its encoded embedding. As a method of self-supervised learning, feature reconstruction aims to help learn expressive node representations, instead of making the encoded representations similar to raw node features. To achieve this target, we found that the decoder design plays an important role in learning meaningful representations. }

\subsection{Training and Inference}
\model is a simple and scalable self-supervised graph learning framework without complex designs and operations.
First, we randomly select a certain proportion of nodes, and replace their node features with the mask-token [MASK]. We feed the partial observed graph into the encoder to generate encoding node representations. Before the decoding, we re-mask the selected nodes and replace them with another token [DMASK]. The decoder is applied to the re-masked graph to reconstruct original node features.

For downstream applications, the encoder is applied to input graph without any masking in the inference stage. The generated node embeddings could be used for downstream tasks, such as node classification. For graph-level tasks, we use a non-parameterized graph pooling (readout) function, i.e., MaxPooling, MeanPooling, to obtain the graph representation
$ \vh^g = \mathrm{READOUT}(\{\vh_i, v_i \in \mathcal{G}_g \}) $.
In the experiments, we show that \model achieve competitive performance in both node-level and graph-level applications. 

The decoder is only used during self-supervised learning to perform node feature reconstruction task. There, the decoder architecture is independent of the encoder design and can use any GNN layer. 

\todo{add a algorithmic pipeline}

}

%% file: 4.experiments.tex
\section{Experiments}%
\label{sec:experiments}

In this section, we demonstrate that \model  is a general self-supervised framework for various graph learning tasks, including:
\begin{itemize}[leftmargin=*,itemsep=0pt,parsep=0.5em,topsep=0.3em,partopsep=0.3em]
    \item Unsupervised representation learning for \textit{node classification};
    \item Unsupervised representation learning for \textit{graph classification};
    \item \textit{Transfer learning} on molecular property prediction.
\end{itemize}
Extensive experiments on various datasets are conducted to evaluate 
the performance of \model against state-of-the-art (SOTA) contrastive and generative methods on these three tasks. 
In each task, we follow exactly the same experimental procedure, e.g., data splits, evaluation protocol, as the  standard settings~\cite{velivckovic2018deep,zhang2021canonical,sun2019infograph,hu2020strategies}. 


\subsection{Node Classification}


\vpara{Setup.}
The node classification task is to predict the unknown node labels in networks. 
We test the performance of \model on 6 standard benchmarks: Cora, Citeseer, PubMed~\cite{yang2016revisiting}, Ogbn-arxiv~\cite{hu2020open}, PPI, and Reddit. 
Following the inductive setup in GraphSage~\cite{hamilton2017inductive}, the testing for Reddit and PPI is carried out on unseen nodes and graphs, while the other networks are used for transductive learning.

For the evaluation protocol, we follow the experimental setting in ~\cite{velivckovic2018deep,hassani2020contrastive,thakoor2021bootstrapped,zhang2021canonical}. 
First, we train a GNN encoder by the proposed \model without supervision. 
Then we freeze the parameters of the encoder and generate all the nodes' embeddings. 
For evaluation, we train a linear classifier and report the mean accuracy on the test nodes through 20 random initializations. 
We follow the public data splits ~\cite{velivckovic2018deep,hassani2020contrastive,zhang2021canonical}  of Cora, Citeseer, and PubMed. 
The graph encoder $f_E$ and decoder $f_D$ are both specified as standard GAT. 
Detailed hyper-parameters can be found in  Appendix \ref{app:appendix}.

\vpara{Results.} 
We compare  \model with SOTA  contrastive self-supervised models, DGI~\cite{velivckovic2018deep}, MVGRL~\cite{hassani2020contrastive}, GRACE~\cite{zhu2020deep}, BGRL~\cite{thakoor2021bootstrapped}, InfoGCL~\cite{Xu2021InfoGCLIG}, and CCA-SSG~\cite{zhang2021canonical}, as well as supervised baselines GCN and GAT. 
We also report the results of previous generative self-supervised models, GAE~\cite{kipf2016variational}, GPT-GNN~\cite{hu2020gpt}, and GATE~\cite{amin2020gate}.
We report results from previous works with the same experimental setup if available.
If results are not previously reported and codes are provided, we implement them based on the official codes and conduct a hyper-parameter search. 
Table~\ref{tab:node_clf} lists the results. 
\model achieves the best or competitive results compared to the SOTA self-supervised approaches in all benchmarks. 
Notably, \model outperforms existing generative methods by a large margin. 
The results in the inductive setting of PPI and Reddit suggest the self-supervised \model technique provides strong generalization to unseen nodes. 



\subsection{Graph Classification}

\vpara{Setup.}
For graph classification,  we conduct experiments on 7 benchmarks: MUTAG, IMDB-B, IMDB-M, PROTEINS, COLLAB, REDDIT-B, and NCI1~\cite{yanardag2015deep}.
Each dataset is a collection of graphs where each graph is associated with a label.
Node labels are used as input features in MUTAG, PROTEINS, and NCI1, whereas node degrees are used in IMDB-B, IMDB-M, REDDIT-B, and COLLAB.

For the evaluation protocol, after generating graph embeddings with \model's encoder and readout function, we feed encoded graph-level representations into a downstream LIBSVM~\cite{chang2011libsvm} classifier to predict the label, and report the mean 10-fold cross-validation accuracy with standard deviation after 5 runs.  We adopt GIN~\cite{xu2019powerful}, which is commonly used in previous graph classification works, as the backbone of encoder and decoder.

\vpara{Results.}
In addition to classical graph kernel methods---Weisfeiler-Lehman sub-tree kernel (WL)~\cite{shervashidze2011weisfeiler} and deep graph kernel (DGK)~\cite{yanardag2015deep}, we also compare \model with SOTA unsupervised and contrastive methods, 
GCC~\cite{qiu2020gcc}, graph2vec~\cite{narayanan2017graph2vec}, Infograph~\cite{sun2019infograph}, GraphCL~\cite{you2020graph}, JOAO~\cite{you2021graph},  MVGRL~\cite{hassani2020contrastive}, and InfoGCL~\cite{Xu2021InfoGCLIG}. 
The supervised baselines, GIN~\cite{xu2019powerful} and DiffPool~\cite{ying2018hierarchical}, are also included. 
Per graph classification research tradition, we report results from previous papers if available.
The results are shown in Table~\ref{tab:graph_clf}.
We find that \model outperforms all self-supervised baselines on five out of seven datasets and has competitive results on the other two.
The node features of these seven datasets are all one-hot vectors representing node-labels or degrees, which are considered to be less informative than node features in node classification. The results manifest that generative auto-encoders could learn meaningful information and offer potentials in graph-level tasks.

\begin{table*}[htbp]
    \centering
    \caption{Experiment results in unsupervised representation learning for \underline{node classification}. \textmd{We report the Micro-F1 (\%) score for PPI and accuracy (\%) for the other datasets. }
    }
    \begin{threeparttable}
    \renewcommand\tabcolsep{10pt}
    \renewcommand\arraystretch{1.05}
    \begin{tabular}{c|c|cccccc}
        \toprule[1.2pt]
            & Dataset &   Cora      & CiteSeer      & PubMed                & Ogbn-arxiv        & PPI               & Reddit        \\

         \midrule
        \multirow{2}{*}{Supervised} 
        & GCN     &  81.5          & 70.3          & 79.0                   & 71.74$\pm$0.29    & 75.7$\pm$0.1    & 95.3$\pm$0.1           \\
        & GAT     &  83.0$\pm$0.7  & 72.5$\pm$0.7  & 79.0$\pm$0.3           & 72.10$\pm$0.13     & 97.30$\pm$0.20    & 96.0$\pm$0.1           \\
        \midrule
        \multirow{10}{*}{Self-supervised} 
        & GAE     &  71.5$\pm$0.4  & 65.8$\pm$0.4  & 72.1$\pm$0.5           & -               & -           & - \\
        & GPT-GNN &  80.1$\pm$1.0  & 68.4$\pm$1.6  & 76.3$\pm$0.8 & - & - & -\\
        & GATE    &  83.2$\pm$0.6  & 71.8$\pm$0.8  & 80.9$\pm$0.3           & -                 & -             & -   \\ 
        & DGI     &  82.3$\pm$0.6  & 71.8$\pm$0.7  & 76.8$\pm$0.6           & 70.34$\pm$0.16 & 63.80$\pm$0.20     & 94.0$\pm$0.10 \\
        & MVGRL   & 83.5$\pm$0.4   & 73.3$\pm$0.5  & 80.1$\pm$0.7           & -               & - & - \\
        & GRACE$^{1}$   & 81.9$\pm$0.4   & 71.2$\pm$0.5  & 80.6$\pm$0.4           & 71.51$\pm$0.11  & 69.71$\pm$0.17  &    94.72$\pm$0.04\\  
        & BGRL$^{1}$    & 82.7$\pm$0.6   & 71.1$\pm$0.8  & 79.6$\pm$0.5           & \underline{71.64$\pm$0.12}   & \underline{73.63$\pm$0.16}  & 94.22$\pm$0.03         \\
        & InfoGCL  & 83.5$\pm$0.3   & \bf 73.5$\pm$0.4  & 79.1$\pm$0.2  & - & - & - \\
        & CCA-SSG$^{1}$ & \underline{84.0$\pm$0.4}   & 73.1$\pm$0.3  & \underline{81.0$\pm$0.4}  & 71.24$\pm$0.20  & 73.34$\pm$0.17  & \underline{95.07$\pm$0.02}   \\
        \cmidrule{2-8}
         & \model  & \bf 84.2±0.4  & \underline{73.4±0.4}  & \bf 81.1±0.4  & \bf 71.75$\pm$0.17 & \bf 74.50$\pm$0.29    & \bf 96.01$\pm$0.08    \\

        \bottomrule[1.2pt]
    \end{tabular}
     \begin{tablenotes}
        \footnotesize
        \item[] The results not reported are due to unavailable code or out-of-memory.
        \item[1] Results are from reproducing using authors' official code, as they did not report the results in part of datasets. The result of PPI is a bit different from what the authors' reported. This is because we train the linear classifier until convergence, rather than for a small fixed number of epochs during evaluation, using the official code.
    \end{tablenotes}

    \end{threeparttable}
    \label{tab:node_clf}
\end{table*}

\begin{table*}[htbp]
    \centering
    \caption{Experiment results in unsupervised representation learning for \underline{graph classification}. \textmd{We report accuracy (\%) for all datasets.}}
    \begin{threeparttable}
    \renewcommand\arraystretch{1.05}
    \begin{tabular}{c|c|ccccccc}
        \toprule[1.2pt]
              & Dataset  & IMDB-B     & IMDB-M     & PROTEINS   & COLLAB     & MUTAG      & REDDIT-B   & NCI1     \\ 

        \midrule
        \multirow{2}{*}{Supervised}
        & GIN         & 75.1$\pm$5.1   & 52.3$\pm$2.8   & 76.2$\pm$2.8   & 80.2$\pm$1.9   & 89.4$\pm$5.6   & 92.4$\pm$2.5   & 82.7$\pm$1.7 \\ 
        & DiffPool    & 72.6$\pm$3.9 &  -           &  75.1$\pm$3.5   & 78.9$\pm$2.3 & 85.0$\pm$10.3 & 92.1$\pm$2.6 & - \\
        \midrule
        \multirow{2}{*}{Graph Kernels}
        & WL          & 72.30$\pm$3.44 & 46.95$\pm$0.46 & 72.92$\pm$0.56 & - & 80.72$\pm$3.00 & 68.82$\pm$0.41 & 80.31$\pm$0.46 \\ 
        & DGK         & 66.96$\pm$0.56 & 44.55$\pm$0.52 & 73.30$\pm$0.82 & - & 87.44$\pm$2.72 & 78.04$\pm$0.39 & 80.31$\pm$0.46 \\ 
        \midrule
        \multirow{8}{*}{Self-supervised}
        & graph2vec   & 71.10$\pm$0.54 & 50.44$\pm$0.87 & 73.30$\pm$2.05 & -              & 83.15$\pm$9.25 & 75.78$\pm$1.03 & 73.22$\pm$1.81 \\ 
        & Infograph   & 73.03$\pm$0.87 & 49.69$\pm$0.53 & 74.44$\pm$0.31 & 70.65$\pm$1.13 & 89.01$\pm$1.13 & 82.50$\pm$1.42 & 76.20$\pm$1.06 \\
        & GraphCL     & 71.14$\pm$0.44 & 48.58$\pm$0.67 & 74.39$\pm$0.45 & 71.36$\pm$1.15 & 86.80$\pm$1.34 & \underline{89.53$\pm$0.84} & 77.87$\pm$0.41 \\
        & JOAO        & 70.21$\pm$3.08 & 49.20$\pm$0.77     & \underline{74.55$\pm$0.41} & 69.50$\pm$0.36 & 87.35$\pm$1.02 & 85.29$\pm$1.35 & 78.07$\pm$0.47 \\
        & GCC         & 72.0           & 49.4           & -    & 78.9    &  - & \bf 89.8 & - \\
        & MVGRL       & 74.20$\pm$0.70   & 51.20$\pm$0.50   & -              & -              & \underline{89.70$\pm$1.10}   & 84.50$\pm$0.60   & -               \\
        & InfoGCL     & \underline{75.10$\pm$0.90}   & \underline{51.40$\pm$0.80}   &  -             & \underline{80.00$\pm$1.30}   & \bf 91.20$\pm$1.30   & -              &  \underline{80.20$\pm$0.60}   \\
        \cmidrule{2-9}
        & \model      & \bf 75.52$\pm$0.66 & \bf 51.63$\pm$0.52 & \bf 75.30$\pm$0.39 & \bf 80.32$\pm$0.46 & 88.19$\pm$1.26 & 88.01$\pm$0.19 & \bf 80.40$\pm$0.30  \\
        \bottomrule[1.2pt]
    \end{tabular}
        \begin{tablenotes}
            \footnotesize
            \item[]   The reported results of baselines are from previous papers if available.
        \end{tablenotes}
    \end{threeparttable}
    \label{tab:graph_clf}
\end{table*}

\begin{table*}[htp]
\caption{Experiment results in \underline{transfer learning} on molecular property prediction benchmarks. \textmd{The model is first pre-trained on ZINC15 and then finetuned on the following datasets. We report ROC-AUC scores (\%).}}
\label{tab:mol_clf}
\renewcommand\tabcolsep{8pt}
\renewcommand\arraystretch{1.05}
\begin{tabular}{c|cccccccc|c}
\toprule[1.2pt]
                        & BBBP       & Tox21      & ToxCast    & SIDER      & ClinTox    & MUV        & HIV        & BACE & Avg.      \\
\midrule
    No-pretrain         & 65.5±1.8   & 74.3±0.5   & 63.3±1.5   & 57.2±0.7   & 58.2±2.8   & 71.7±2.3   & 75.4±1.5   & 70.0±2.5 & 67.0  \\
\midrule    
    ContextPred         & 64.3±2.8   & \underline{75.7±0.7}   & 63.9±0.6   & 60.9±0.6   & 65.9±3.8   & 75.8±1.7   & 77.3±1.0   & 79.6±1.2 & 70.4 \\
    AttrMasking         & 64.3±2.8   & \bf 76.7±0.4   & \bf 64.2±0.5   &  \underline{61.0±0.7}   & 71.8±4.1   & 74.7±1.4   & 77.2±1.1   & 79.3±1.6 & 71.1 \\
    Infomax             & 68.8 ±0.8  & 75.3 ±0.5  & 62.7 ±0.4  & 58.4 ±0.8  & 69.9±3.0  & 75.3 ±2.5  & 76.0 ±0.7  & 75.9 ±1.6 & 70.3 \\
    GraphCL             & 69.7±0.7 & 73.9±0.7 & 62.4±0.6 & 60.5±0.9 & 76.0±2.7 & 69.8±2.7 & \bf 78.5±1.2 & 75.4±1.4 & 70.8 \\
    JOAO                & 70.2±1.0 & 75.0±0.3 & 62.9±0.5 & 60.0±0.8 & \underline{81.3±2.5} & 71.7±1.4 & 76.7±1.2 & 77.3±0.5 & 71.9 \\
    GraphLoG            & \bf 72.5±0.8 &  \underline{75.7±0.5}  &  63.5±0.7     & \bf 61.2±1.1  & 76.7±3.3   & \underline{76.0±1.1}    & \underline{77.8±0.8}  & \bf 83.5±1.2 & \underline{73.4} \\ 
\midrule
    GraphMAE            & \underline{72.0±0.6} & 75.5±0.6 & \underline{64.1±0.3} & 60.3±1.1 & \bf 82.3±1.2 & \bf 76.3±2.4 & 77.2±1.0 & \underline{83.1±0.9} & \bf 73.8 \\
\bottomrule[1.2pt]
\end{tabular}
\end{table*}

\subsection{Transfer Learning}

\vpara{Setup.} To evaluate the transferability of the proposed method, we test the performance on transfer learning on molecular property prediction, following the setting of ~\cite{hu2020strategies,you2020graph,you2021graph}. The model is first pre-trained in  2 million unlabeled molecules sampled from the ZINC15~\cite{sterling2015zinc}, and then finetuned in 8 classification benchmark datasets contained in MoleculeNet~\cite{wu2018moleculenet}. The downstream datasets are split by scaffold-split to mimic real-world use cases. Input node features are the atom number and chirality tag, and edge features are the bond type and direction. 

For the evaluation protocol,  we run experiments for 10 times and report the mean and standard deviation of ROC-AUC scores (\%). Following the default setting in ~\cite{hu2020strategies},  we adopt a 5-layer GIN as the encoder and a single-layer GIN as the decoder.

\vpara{Results.}
We evaluate \model against methods including Infomax, AttrMasking and ContextPred~\cite{hu2020strategies} , and SOTA contrastive learning methods, GraphCL~\cite{you2020graph}, JOAO~\cite{you2021graph}, and GraphLoG~\cite{xu2021self}. Table \ref{tab:graph_clf} shows that the performance on downstream tasks is comparable to SOTA methods, in which \model achieves the best average scores and has a small edge over previous best results in two tasks. 
This demonstrates the robust transferability of \model.

To summarize,  the self-supervised \model  method achieves  competitive performance on node classification, graph classification, and transfer learning across 21 benchmarks. 
Note that we do not customize a dedicated \model for each task. 
The consistent results on the three tasks demonstrate that \model is an effective and universal self-supervised graph pre-training framework for various applications.


\hide{
\begin{figure}
    \centering
    \includegraphics[width=0.43\textwidth]{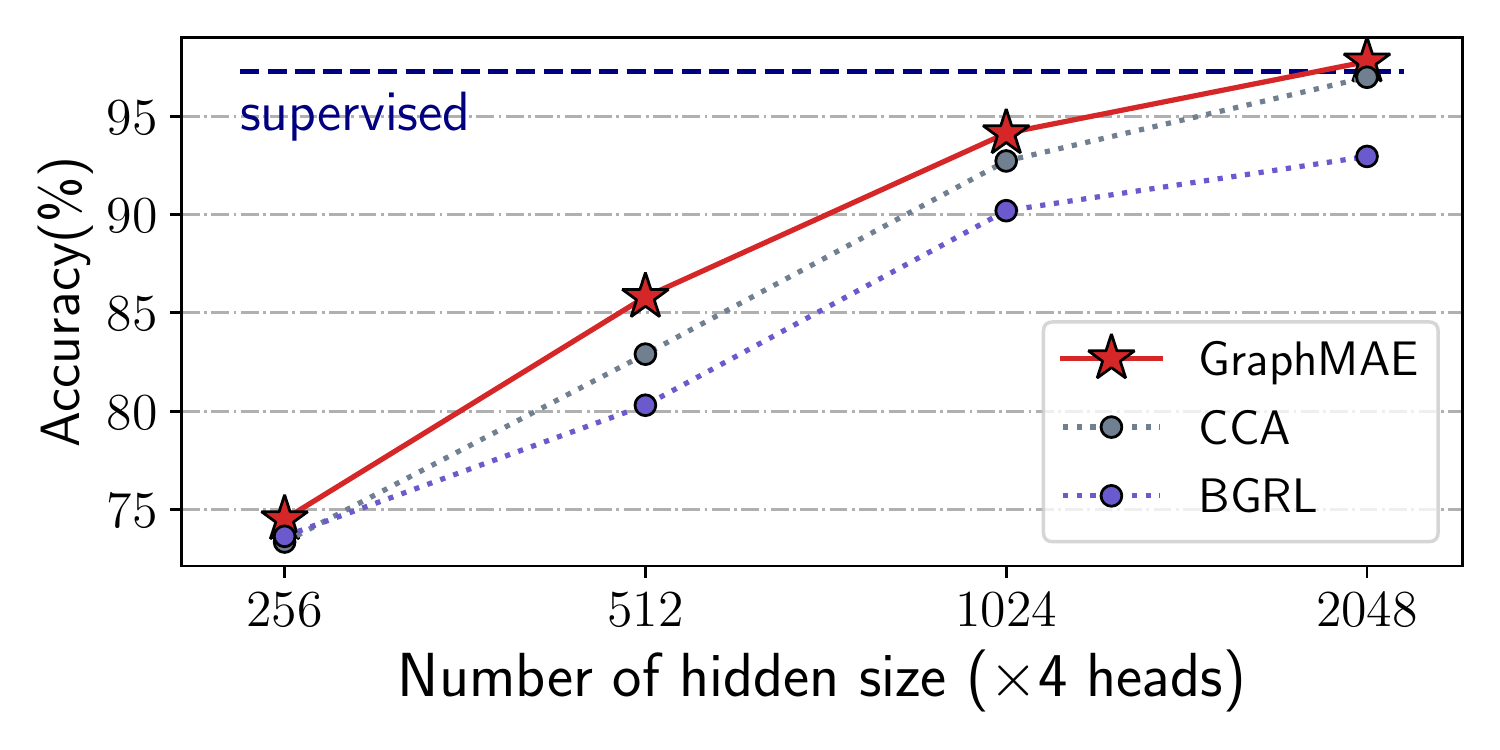}
    \vspace{-4mm}
    \caption{Performance on PPI using GAT with 4 attention heads, compared to other baselines. Self-supervised methods benefit much from larger model size, and \model could outperform supervised model. }
    \label{fig:ppi_hidden}
\end{figure}
}

\subsection{Ablation Studies}
To verify the effects of the main components in \model, we further conduct several ablation studies. Without loss of generalization, we choose three datasets from node classification and two datasets from graph classification for experiments.

\vpara{Effect of reconstruction criterion.} 
We study the influence of reconstruction criterion, and Table \ref{tab:ablation} shows the results of MSE and the SCE loss function.
Generally, the input features in node classification lie in continuous or high-dimensional discrete space, containing more discriminative information.
The results manifest that SCE has a significant advantage over MSE,
with an absolute gain of 1.5\% $\sim$ 8.0\%.
In graph classification, pre-training with either MSE or SCE improves accuracy. 
The input features 
are discrete one-hot encoding in these benchmarks, 
representing either node degrees or node labels. 
Reconstructing one-hot encoding with MSE is
similar to classification tasks, thus MSE also works. Nevertheless, SCE offers a performance edge (though limited) over MSE. 

Figure \ref{fig:ablation} shows the influence of scaling factor $\gamma$.  We observe that $\gamma > 1$ offers benefits in most cases, especially in node classification. However, in MUTAG, a larger $\gamma$ value harms the performance. In our experiments, we notice that the training loss in node classification is much higher than that in graph classification. 
This further demonstrates that aligning continuous vectors in a unit sphere is more challenging. Therefore, scaling $\gamma$ brings improvements.

\begin{table}[t]
\caption{Ablation studies of the decoder type, re-mask and reconstruction criterion on node- and graph-level datasets.}
\label{tab:ablation}
\renewcommand\tabcolsep{4pt}
\input{tables/ablation}

\vspace{-2mm}
\end{table}

\vpara{Effect of mask and re-mask.} Masking plays an important role in the proposed \model method. \model employs two masking strategies --- masking input feature before encoder, and re-masking encoded code before decoder. Table \ref{tab:ablation} studies the designs. We observe a significant drop in performance if not masking input features, indicating that masking inputs is vital to avoid  the trivial solution. For the re-mask strategy, the accuracy drops by 0.1\%$\sim$1.9\% without it. 
Re-mask is designed for the GNN decoder and can be regarded as regularization, which makes the self-supervised task more challenging.

\hide{
\begin{table}
\centering
\caption{Experiment results using different encoder backbones in node classification.}
\renewcommand\tabcolsep{3pt}
\begin{tabular}{c|cccc}
\toprule[1.2pt]
          & Cora & Citeseer & Pubmed & Ogbn-arxiv \\
\midrule
BGRL (GCN)    &   82.7$\pm$0.6    &  71.1$\pm$0.7        & 79.4$\pm$0.6       &  71.64$\pm$0.12      \\
BGRL (GAT)    &   82.8$\pm$0.5    &   71.1$\pm$0.8       &  79.6$\pm$0.5      &   70.07$\pm$0.02 \\
CCA-SSG (GCN) &   84.0$\pm$0.4      &  73.1$\pm$0.3         &  81.0$\pm$0.4        &  70.81$\pm$0.13  \\
CCA-SSG (GAT) &   83.8$\pm$0.5    &  72.6$\pm$0.7        &   79.9$\pm$1.1     &  71.24$\pm$0.20  \\
\model (GCN)  &   82.9$\pm$0.6             &  72.5$\pm$0.5                 &  81.0$\pm$0.5               &  \bf 71.87$\pm$0.21\\   
\model (GAT)  &   \bf 84.2$\pm$0.4        &  \bf 73.4$\pm$0.4       & \bf 81.1$\pm$0.4      &  71.75$\pm$0.17 \\
\bottomrule[1.2pt]
\end{tabular}
\vspace{-2mm}
\label{tab:backbone}
\end{table}
}

\begin{figure}
    \centering
    \begin{minipage}[t]{0.235\textwidth}
        \includegraphics[width=\textwidth]{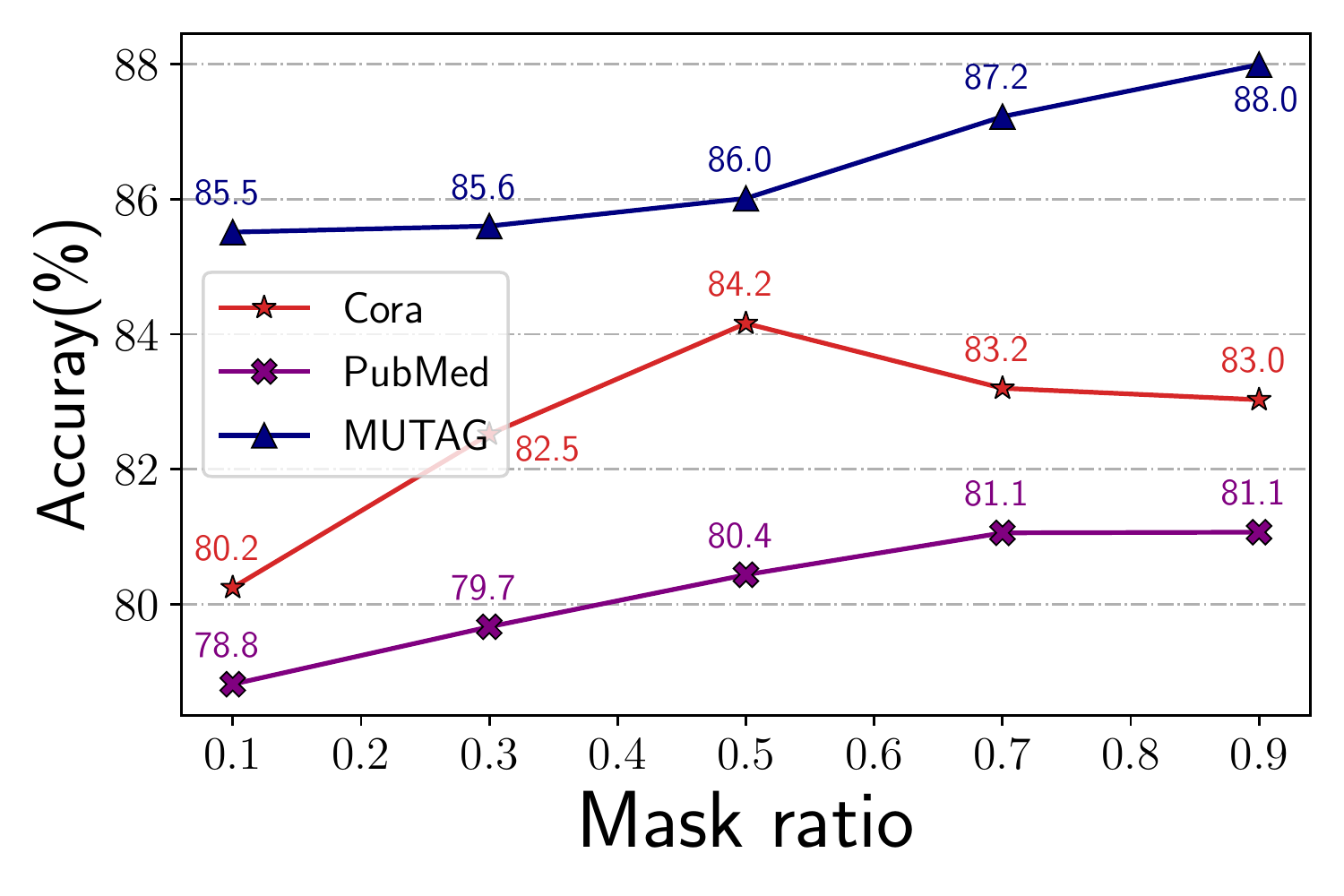}
    \end{minipage}
    \begin{minipage}[t]{0.235\textwidth}
        \includegraphics[width=\textwidth]{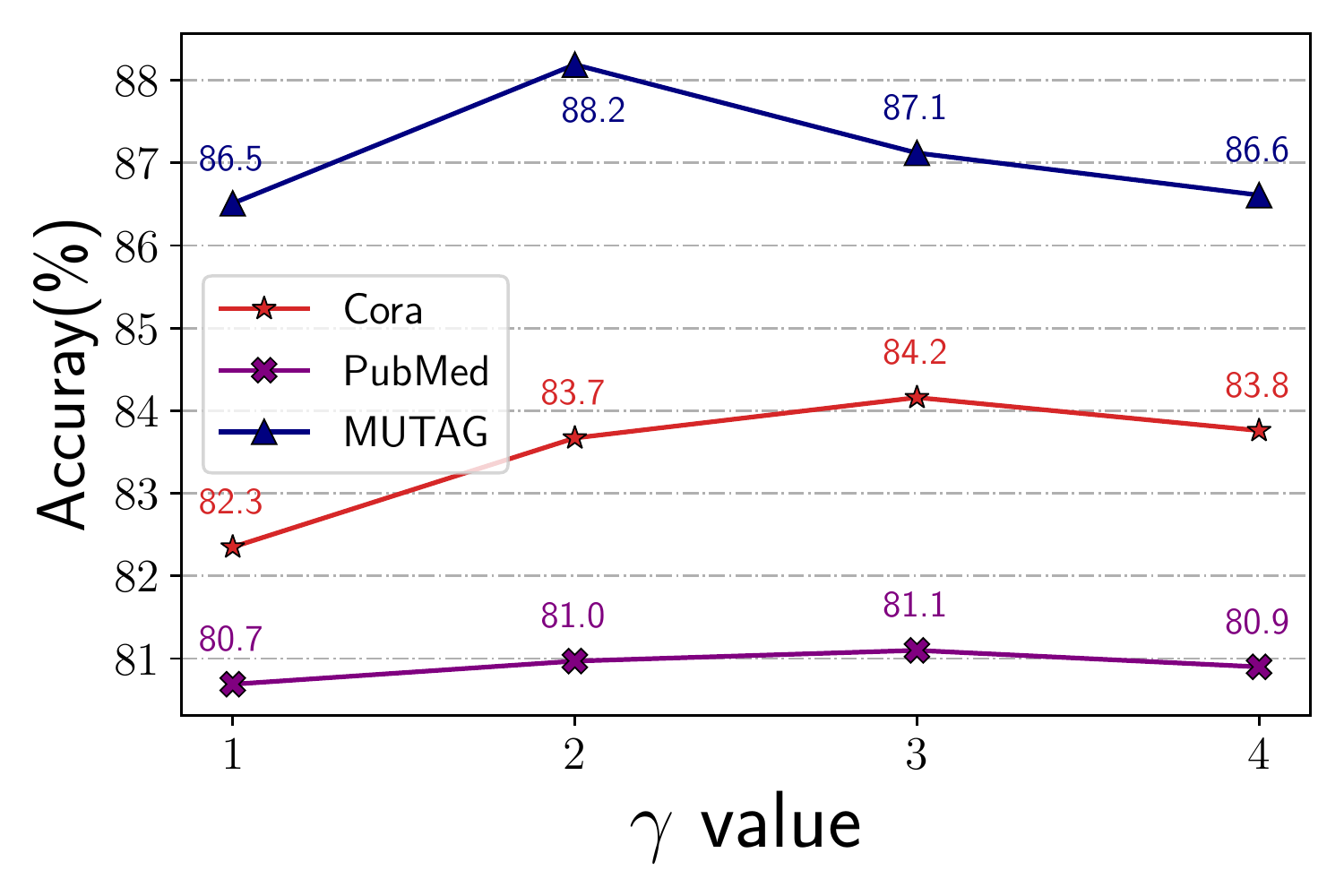}
    \end{minipage}
    \vspace{-6mm}
    \caption{Ablation studies of mask ratio and scaling factor.}
    \label{fig:ablation}
    \vspace{-4mm}
\end{figure}

\vpara{Effect of mask ratio.} 
Figure \ref{fig:ablation} shows the influence of mask ratio. 
In most cases, the reconstruction task with a low mask ratio (0.1) is not challenging enough to learn useful features.
The optimal ratio varies across graphs. 
Results on Ogbn-arxiv and IMDB-B can be found in Appendix \ref{app:appendix}. 
In Cora, increasing the mask ratio by more than 0.5 degrades the performance, while \model still works with a surprisingly high ratio (0.7$\sim$0.9) in PubMed and MUTAG. 
This can be connected with the information redundancy in graphs. 
Large node degrees or high homogeneity may lead to heavy information redundancy, in which missing node features may be recovered from very few neighboring nodes with little high-level understanding of
features and local context. 
In contrast, lower redundancy means that an excessively high mask ratio would make it impossible to recover features, thus degrading the performance. 

\vpara{Effect of decoder type.} 
In Table \ref{tab:ablation}, we compare different decoder types, including MLP, GCN, GIN, and GAT. The re-mask strategy is only used for GNN decoders. As the results show, using GNN decoder typically boosts the performance. Compared to MLP, which reconstructs original features from latent representations, GNN enforces masked nodes to extract information relevant to their original features from the neighborhood. One reasonable assumption is that GNN avoids the representation tending to be the same as original features.
MLP also works in \model, which might be partly attribute to the usage of the SCE metric.

Among different GNNs, GIN performs better in graph-level tasks and GAT is a more reasonable option for node classification. It is observed that replacing GAT with GCN causes a significant drop, especially in Cora ($\sim$2.9\%) and PubMed ($\sim$2.0\%). 
We speculate that the attention mechanism matters in reconstructing continuous features with the re-mask strategy.

\hide{
\vpara{Effect of encoder architecture.}  In node classification, \model uses GAT as the encoder. To have a fair comparison and investigate the influence of different GNN backbones, we compare the best baselines, BGRL and CCA-SSG, in node classification datasets using GCN and GAT as the encoder. The results are shown in Table \ref{tab:backbone}. We observe that \model still outperforms the baselines with the same GAT backbone. 
In addition, the results manifest that attention mechanism would not always benefit graph self-supervised learning, as GCN is inferior to GAT, for instance, in (Cora, CCA-SSG) and (Ogbn-arxiv, \model). Under the training setting of \model, GAT could be a better option in most cases.
}

\hide{

\section{Experiments}%
\label{sec:experiments}

In this section, we show our proposed \model is a general self-supervised framework for a wide range of graph tasks, including:
\begin{itemize}[leftmargin=*,itemsep=0pt,parsep=0.5em,topsep=0.3em,partopsep=0.3em]
    \item Unsupervised representation learning for \textit{node classification}
    \item Unsupervised representation learning for \textit{graph classification}
    \item \textit{Transfer learning} on molecular property prediction.
\end{itemize}
\noindent We comprehensively evaluate the performance of \model against state-of-the-art methods on these three tasks. 

\subsection{Task Evalution}

\subsubsection{Node classification}\hfill

\vpara{Setup.}
The node classification task is to predict unknown node labels in  network. We analyze the performance of \model on a set of 6 standard transductive and inductive benchmarks. Cora, Citeseer, PubMed, and Ogbn-arxiv are citation networks where nodes correspond to documents and edges
represent citations. 
PPI is a protein-protein interaction dataset and  Reddit contains posts belonging to different communities with user comments. 
Following the inductive setup in ~\cite{hamilton2017inductive}, the testing for Reddit and PPI is carried out on unseen (untrained) nodes and graphs, while the citation networks are used for transductive learning.

For evaluation protocol, we follow experiment settings in ~\cite{velivckovic2018deep,hassani2020contrastive,thakoor2021bootstrapped,zhang2021canonical}. We first train the model using the method proposed in this paper, without supervision. Then we freeze the parameters of the encoder and obtain all the nodes' embeddings. For the evaluation, we train a linear classifier and report the mean classification accuracy with standard deviation on the test nodes through 20 random initialization. We follow the public data split ~\cite{velivckovic2018deep,hassani2020contrastive,zhang2021canonical}  of Cora/Citeseer/PubMed. The graph encoder $f_E(·|\theta)$ is specified as a standard GAT model and the decoder $f_D(·|\phi)$ is a single-layer GAT. To have a fair comparison, we also implement previous SOTA contrastive methods based on GAT and conduct hyper-parameter search, as shown in Table \ref{tab:node_clf}. GAT doesn't always outperform GCN, indicating that GAT is not a absolute winner in self-supervised learning. Under the training setting of \model, GAT could be a better option.
Detailed hyper-parameters can be found in the Appendix.

\vpara{Experiment results.} 
We compare our \model with state-of-the-art (SOTA) contrastive self-supervised models, DGI~\cite{velivckovic2018deep}, MVGRL~\cite{hassani2020contrastive}, GRACE~\cite{zhu2020deep}, BGRL~\cite{thakoor2021bootstrapped}, and CCA-SSG~\cite{zhang2021canonical}, and supervised baselines GCN and GAT. We also report the results of previous generative self-supervised models, GAE~\cite{kipf2016variational}, GPT-GNN~\cite{hu2020gpt}, and GATE~\cite{amin2020gate}.
We report results from previous papers with the same experimental setup if available. If results are not previously reported and codes are provided, we implement them based on the official codes and conduct a hyper-parameter search. Otherwise we only report the results on Cora/Citeseer/PubMed using our re-implemented codes according to the paper. Table~\ref{tab:node_clf} represents the results, and it shows that our approach achieves the best or competitive results compared to the state-of-the-art self-supervised approaches. To test the influence of different GNN backbones, we also compare the best baselines, BGRL and CCA-SSG, in citation networks using GCN and GAT as the backbone. The results is shown in Table \ref{tab:backbone}. On the one hand, we observe that \model still outperforms the baselines with the same GAT backbone. 
On the other hand, the results manifest that, attention mechanism would not always bring benefits in graph self-supervised learning, as GCN is inferior to GAT, for instance, in (Cora, CCA-SSG) and (Ogbn-arxiv, \model).
Besides, the results in the inductive setting of PPI and Reddit substantiates the ability of generalization to unseen nodes. \model successfully outperformed all the competing self-supervised methods.  
Previous studies often reports a large gap between supervised and self-supervised results on PPI. We found that increasing the number of model parameters helps little in supervised setting, but could highly boost the performance in self-supervised setting. As show in Table~\ref{tab:ppi_hd}, \model even outperforms supervised counter-part, although the model would be too much larger. This might indicate the potential of large-scale GNNs in pre-training.


\begin{table}
\centering
\small
\caption{Experiment results using different encoder backbones in node classification.}
\renewcommand\tabcolsep{3pt}
\begin{tabular}{c|cccc}
\toprule
          & Cora & Citeseer & Pubmed & Ogbn-arxiv \\
\midrule
BGRL (GCN)    &   82.66$\pm$0.60    &  71.06$\pm$0.66        & 79.44$\pm$0.58       &  71.64$\pm$0.12      \\
BGRL (GAT)    &   82.75$\pm$0.52    &   71.08$\pm$0.79       &  79.55$\pm$0.50      &   70.07$\pm$0.02 \\
CCA-SSG (GCN) &   84.0$\pm$0.4      &  73.1$\pm$0.3          &  80.9$\pm$0.4        &  70.81$\pm$0.13  \\
CCA-SSG (GAT) &   83.78$\pm$0.48    &  72.59$\pm$0.66        &   79.91$\pm$1.09     &  71.24$\pm$0.20  \\
\model (GCN)  &   82.91$\pm$0.56             &  72.50$\pm$0.45                 &  81.05$\pm$0.54               &  \bf 71.87$\pm$0.21\\   
\model (GAT)  &   \bf 84.16±0.44        &  \bf 73.35$\pm$0.42       & \bf 81.10$\pm$0.41      &  71.75$\pm$0.17 \\
\bottomrule
\end{tabular}
\label{tab:backbone}
\end{table}

\subsubsection{Graph classification}\hfill

\vpara{Setup.}
For unsupervised learning for graph classification,  we conduct experiments on 8 well-known benchmarks: MUTAG, IMDB-B, IMDB-M, PROTEINS, COLLAB, REDDIT-B, DD, and NCI1~\cite{yanardag2015deep}, which are widely used in recent graph classification models. Each dataset is a set of graphs where each graph is associated with a label. MUTAG, PROTEINS, DD, and NCI1 have node labels as input features, while IMDB-B, IMDB-M, REDDIT-B, and COLLAB use node degrees as input features. 

For evaluation protocol, after generating graph embeddings with \model's encoder, we feed encoded graph-level representation into a down-stream LIBSVM~\cite{chang2011libsvm} classifier to predict the label, and report the mean 10-fold cross validation accuracy with standard deviation after 5 runs.  We adopt GIN~\cite{xu2019powerful}, which is commonly used in previous graph classification works, as the backbone of encoder and decoder.

\vpara{Experiment results.}
Aside from classical graph kernel methods, Weisfeiler-Lehman sub-tree kernel (WL)~\cite{shervashidze2011weisfeiler} and deep graph kernel (DGK)~\cite{yanardag2015deep}, we also compare \model with SOTA unsupervised and contrastive learning methods, graph2vec~\cite{narayanan2017graph2vec}, Infograph~\cite{sun2019infograph}, GraphCL~\cite{you2020graph}, JOAO~\cite{you2021graph}, and InfoGCL~\cite{Xu2021InfoGCLIG}. The performance of supervised baselines, GIN~\cite{xu2019powerful} and DGCNN~\cite{Zhang2018AnED} is also included. We report results from previous papers if available.
The results are shown in Table~\ref{tab:graph_clf}.
We find that \model outperforms all unsupervised baselines on 4 out of 7 of the datasets, and has competitive results in two of the other. In these benchmarks, node features are all one-hot vectors, representing node-labels or degrees, which are often considered to be less informative compared with node features in node classification.  The results manifest that generative auto-encoding could also learn meaningful information and be potentional in graph-level tasks.

\begin{table*}[htbp]
    \centering
    \caption{\textmd{Experiment results in unsupervised representation learning for \textbf{\textit{node classification}}. We report Micro-F1(\%) score for PPI and accuracy(\%) for the other datasets. }}
    \begin{threeparttable}
    \renewcommand\tabcolsep{8pt}
    \begin{tabular}{c|c|cccccc}
        \toprule[1.2pt]
            & Dataset &   Cora      & CiteSeer      & PubMed                & Ogbn-arxiv        & PPI               & Reddit        \\

         \midrule
        \multirow{2}{*}{Supervised} 
        & GCN     &  81.5          & 70.3          & 79.0                   & 71.74$\pm$0.29    & 75.7$\pm$0.1    & 95.4           \\
        & GAT     &  83.0$\pm$0.7  & 72.5$\pm$0.7  & 79.0$\pm$0.3           & 72.1$\pm$0.13     & 97.30$\pm$0.20    & 96.5           \\
        \midrule
        \multirow{10}{*}{Self-supervised} 
        & GAE     &  71.5$\pm$0.4  & 65.8$\pm$0.4  & 72.1$\pm$0.5           & OOM               & OOM           & OOM \\
        & GPT-GNN &  80.1$\pm$1.0  & 68.4$\pm$1.6  & 76.3$\pm$0.8 & - & - & -\\
        & GATE    &  83.2$\pm$0.6  & 71.8$\pm$0.8  & \underline{80.9$\pm$0.3}           & -                 & -             & -   \\ 
        & DGI     &  82.3$\pm$0.6  & 71.8$\pm$0.7  & 76.8$\pm$0.6           & 70.34$\pm$0.16 & 63.8$\pm$0.20     & 94.0$\pm$0.10 \\
        & MVGRL   & 83.5$\pm$0.4   & 73.3$\pm$0.5  & 80.1$\pm$0.7           & OOM               & OOM & OOM \\
        & GRACE$^{1}$   & 81.9$\pm$0.4   & 71.2$\pm$0.5  & 80.6$\pm$0.4           & 71.51$\pm$0.11  & 69.71$\pm$0.17  &    94.72$\pm$0.04\\  
        & BGRL$^{1}$    & 82.7$\pm$0.6   & 71.1$\pm$0.8  & 79.6$\pm$0.5           & \underline{71.64$\pm$0.12}   & \underline{73.63$\pm$0.16}  & 94.22$\pm$0.03         \\
        & InfoGCL$^2$  & 83.5$\pm$0.3   & \bf 73.5$\pm$0.4  & 79.1$\pm$0.2  & - & - & - \\
        & CCA-SSG$^{1}$ & \underline{84.0$\pm$0.4}   & 73.1$\pm$0.3  & \underline{80.9$\pm$0.4}  & 71.24$\pm$0.20  & 73.34$\pm$0.17  & \underline{95.07$\pm$0.02}   \\
        \cmidrule{2-8}
         & \model  & \bf 84.16$\pm$0.44  & \underline{73.35$\pm$0.42}  & \bf 81.10$\pm$0.41  & \bf 71.75$\pm$0.17 & \bf 74.50$\pm$0.29    & \bf 96.01$\pm$0.08    \\
        \bottomrule[1.2pt]
    \end{tabular}
     \begin{tablenotes}
        \footnotesize
        \item[1] Results are from our reproducing with authors’ public code, as they didn't report the results in part of these datasets. The result of PPI is a bit different from that the authers' reported. This is because we use the authors' public code, and train the linear classifier until convergence, instead for a fixed number of epochs, in downstream evaluation.
        \item[2] The code is not publicly available.
    \end{tablenotes}

    \end{threeparttable}
    \label{tab:node_clf}
\end{table*}

\begin{table*}[htbp]
    \centering
    \caption{\textmd{Experiment results in unsupervised representation learning for \textbf{\textit{graph classification}}. We report accuracy(\%) for all datasets.}}
    \begin{threeparttable}
    \begin{tabular}{c|c|ccccccc}
        \toprule[1.2pt]
              & Dataset  & IMDB-B     & IMDB-M     & PROTEINS   & COLLAB     & MUTAG      & REDDIT-B   & NCI1     \\ 

        \midrule
        \multirow{2}{*}{Supervised}
        & GIN         & 75.1$\pm$5.1   & 52.3$\pm$2.8   & 76.2$\pm$2.8   & 80.2$\pm$1.9   & 89.4$\pm$5.6   & 92.4$\pm$2.5   & 82.7$\pm$1.7 \\ 
        & DiffPool \\
        \midrule
        \multirow{2}{*}{Graph Kernels}
        & WL          & 72.30$\pm$3.44 & 46.95$\pm$0.46 & 72.92$\pm$0.56 & - & 80.72$\pm$3.00 & 68.82$\pm$0.41 & 80.31$\pm$0.46 \\ 
        & DGK         & 66.96$\pm$0.56 & 44.55$\pm$0.52 & 73.30$\pm$0.82 & - & 87.44$\pm$2.72 & 78.04$\pm$0.39 & 80.31$\pm$0.46 \\ 
        \midrule
        \multirow{8}{*}{Self-supervised}
        & graph2vec   & 71.10$\pm$0.54 & 50.44$\pm$0.87 & 73.30$\pm$2.05 & -              & 83.15$\pm$9.25 & 75.78$\pm$1.03 & 73.22$\pm$1.81 \\ 
        & Infograph   & 73.03$\pm$0.87 & 49.69$\pm$0.53 & 74.44$\pm$0.31 & 70.65$\pm$1.13 & 89.01$\pm$1.13 & 82.50$\pm$1.42 & 76.20$\pm$1.06 \\
        & GraphCL     & 71.14$\pm$0.44 & 48.58$\pm$0.67 & 74.39$\pm$0.45 & 71.36$\pm$1.15 & 86.80$\pm$1.34 & \underline{89.53$\pm$0.84} & 77.87$\pm$0.41 \\
        & JOAO        & 70.21$\pm$3.08 & 49.20$\pm$0.77     & \underline{74.55$\pm$0.41} & 69.50$\pm$0.36 & 87.35$\pm$1.02 & 85.29$\pm$1.35 & 78.07$\pm$0.47 \\
        & GCC         & 72.0           & 49.4           & -    & 78.9    &  - & \bf 89.8 & - \\
        & MVGRL       & 74.2$\pm$0.7   & 51.2$\pm$0.5   & -              & -              & \underline{89.7$\pm$1.1}   & 84.5$\pm$0.6   & -               \\
        & InfoGCL     & \underline{75.1$\pm$0.9}   & \underline{51.4$\pm$0.8}   &  -             & \underline{80.0$\pm$1.3}   & \bf 91.2$\pm$1.3   & -              &  \underline{80.2$\pm$0.6}   \\
        \cmidrule{2-9}
        & \model      & \bf 75.52$\pm$0.66 & \bf 51.63$\pm$0.52 & \bf 75.30$\pm$0.39 & \bf 80.32$\pm$0.46 & 88.19$\pm$1.26 & 88.01$\pm$0.19 & \bf 80.40$\pm$0.30  \\
        \bottomrule[1.2pt]
    \end{tabular}
        \begin{tablenotes}
            \footnotesize
            \item[]   The reported results of baselines are from previous papers if available.
        \end{tablenotes}
    \end{threeparttable}
    \label{tab:graph_clf}
\end{table*}

\begin{table*}[htp]
\caption{\textmd{Experiment results in \textbf{\textit{transfer learning}} on molecular property prediction benchmarks. The model is first pre-trained on ZINC15 and then finetuned on the following datasets. We report ROC-AUC(\%) scores.}}
\label{tab:mol_clf}
\renewcommand\tabcolsep{6pt}
\begin{tabular}{c|cccccccc}
\toprule[1.2pt]
                        & BBBP       & Tox21      & ToxCast    & SIDER      & ClinTox    & MUV        & HIV        & BACE       \\
\midrule
    No-pretrain         & 65.5$\pm$1.8   & 74.3$\pm$0.5   & 63.3$\pm$1.5   & 57.2$\pm$0.7   & 58.2$\pm$2.8   & 71.7$\pm$2.3   & 75.4$\pm$1.5   & 70.0$\pm$2.5   \\
    ContextPred         & 64.3$\pm$2.8   & 75.7$\pm$0.7   & 63.9$\pm$0.6   & 60.9$\pm$0.6   & 65.9$\pm$3.8   & 75.8$\pm$1.7   & 77.3$\pm$1.0   & 79.6$\pm$1.2   \\
    AttrMasking         & 64.3$\pm$2.8   & 76.7$\pm$0.4   & 64.2$\pm$0.5   & 61.0$\pm$0.7   & 71.8$\pm$4.1   & 74.7$\pm$1.4   & 77.2$\pm$1.1   & 79.3$\pm$1.6   \\
    Infomax             & 68.8$\pm$0.8  & 75.3$\pm$0.5  & 62.7$\pm$0.4  & 58.4 $\pm$0.8  & 69.9$\pm$3.0  & 75.3$\pm$2.5  & 76.0$\pm$0.7  & 75.9$\pm$1.6  \\
    GraphCL             & 69.68$\pm$0.67 & 73.87$\pm$0.66 & 62.40$\pm$0.57 & 60.53$\pm$0.88 & 75.99$\pm$2.65 & 69.80$\pm$2.66 & 78.47$\pm$1.22 & 75.38$\pm$1.44 \\
    JOAO                & 70.22$\pm$0.98 & 74.98$\pm$0.29 & 62.94$\pm$0.48 & 59.97$\pm$0.79 & 81.32$\pm$2.49 & 71.66$\pm$1.43 & 76.73$\pm$1.23 & 77.34$\pm$0.48 \\
    GraphLog \\ 
\midrule
    GraphMAE        & 72.04$\pm$0.66 & 75.51$\pm$0.61 & 64.06$\pm$0.29 & 60.25$\pm$1.13 & 82.32$\pm$1.15 & 76.26$\pm$2.39 & 77.19$\pm$0.95 & 83.13$\pm$0.86 \\
\bottomrule[1.2pt]
\end{tabular}
\end{table*}

\subsubsection{Transfer Learning}\hfill

\vpara{Setup.} To evaluate the transferability of our proposed method, we test the performance on transfer learning on molecular property prediction in chemistry, following the setting of ~\cite{hu2020strategies,you2020graph,you2021graph}. The model is first pre-trained in  2 million unlabeled molecules sampled from the ZINC15 database~\cite{sterling2015zinc}, and then finetuned in 8 binary classification benchmark datasets contained in MoleculeNet~\cite{wu2018moleculenet}. The downstream datasets are splitted by scaffold-split to mimic real-world use case.  Atom number and chirality tag are input as node features and bond type and direction are regarded as edge features. In our experiments, we only consider reconstructing node features.

For evaluation protocal,  we run experiments for 10 times and report the mean and standard deviation of ROC-AUC scores(\%). Following the default setting in ~\cite{hu2020strategies},  we adopt a 5-layer GIN as backbone of the encoder, and a single-layer GIN as the decoder.

\vpara{Experiment results.}
We evaluate the performance of \model against SOTA self-supervised pre-training methods for GNNs, AttrMasking~\cite{hu2020strategies} and ContextPred~\cite{hu2020strategies} , and SOTA contrastive learning methods, GraphCL~\cite{you2020graph}, JOAO~\cite{you2021graph}, and Grover~\cite{rong2020self}. We also report the result of another generating pre-training method, GPT-GNN~\cite{hu2020gpt}. Table \ref{tab:graph_clf} shows that the performance on downstream molecular property prediction tasks is comparable to SOTA self-supervised methods, in which our model has a small edge over previous best results in 4 datasets.

To summarize, our \model achieves the competitive performance on node, graph classification, and transfer learning benchmarks using a unified learning approach. and we do not devise a specialized self-supervised learning method for each task. The results in three tasks demonstrate that \model is effective and universal approach for various applications.


\begin{figure}
    \centering
    \includegraphics[width=0.45\textwidth]{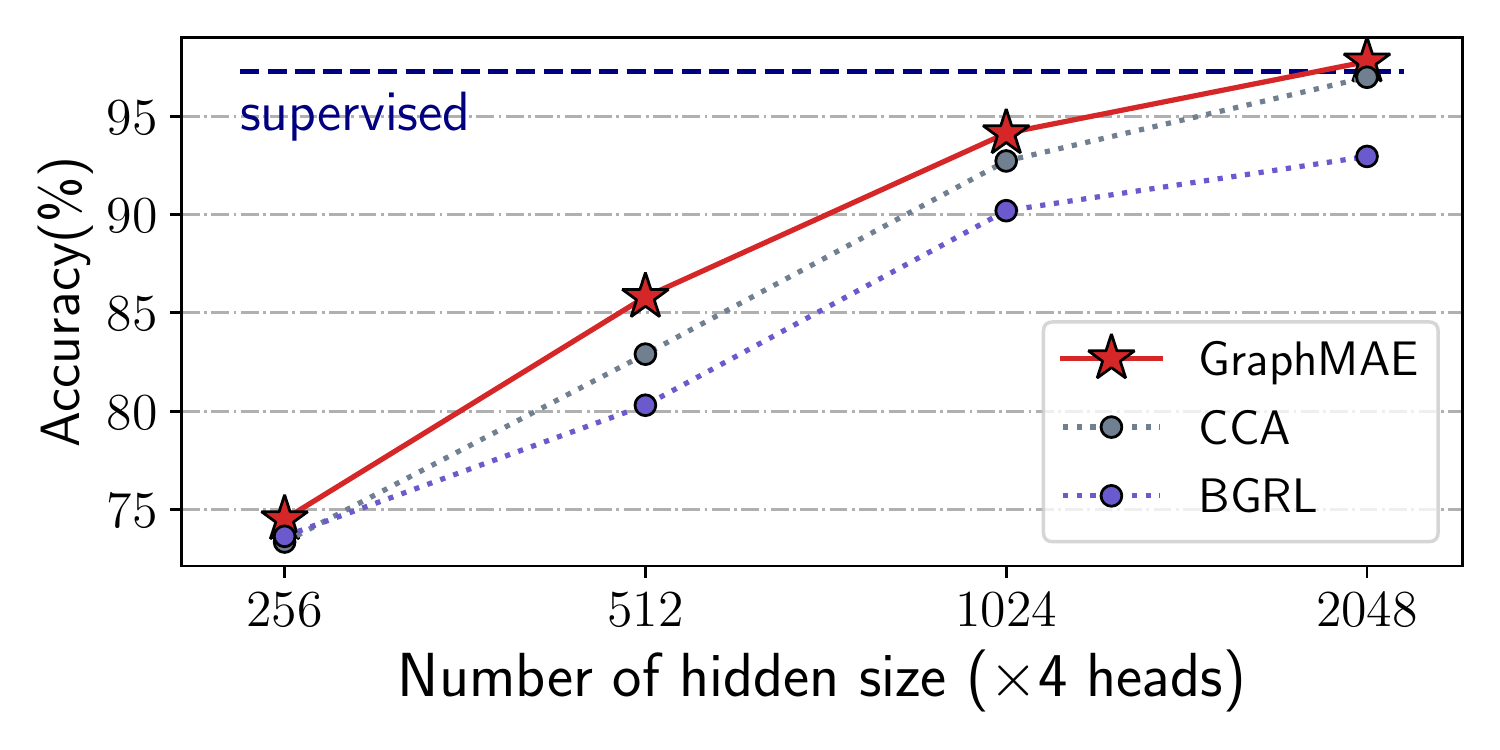}
    \vspace{-4mm}
    \caption{Performance on PPI using GAT with 4 attention heads, compared to other baselines. Self-supervised methods benefit much from larger model size, and \model could outperform supervised model. }
    \label{fig:ppi_hidden}
\end{figure}

\subsection{Ablation Studies}
To verify the effects of main components of our model, we further conduct a series of ablation studies. Without loss of generalization, we choose 3 datasets from node classification and 2 datasets from graph classification as the study objects.

\vpara{Effect of reconstruction criterion.} We study the influence of reconstruction criterion, and Table \ref{tab:ablation} shows the results of MSE and our proposed SCE loss function. Node features are normalized in preprocessing. This helps reconstruction and improves performance in downstream evaluation. The conclusion from Table \ref{tab:ablation} varies between node classification and graph classification.
Generally, input features in node classification lie in continuous space, containing more discriminative information.
The results manifest that SCE has significant advantage over MSE as replacing MSE with SCE brings 4.1\% $\sim$ 15.6\% absolute gain.
In graph classification, pre-training with either MSE or SCE as criterion improves accuracy. The input features typically lie in discrete value space, and are one-hot encoding in these benchmarks, representing degree or node-label. Reconstructing one-hot feature is, to some extent, similar to a classification task, and MSE works in this setting of discrete node feature.  Nevertheless, SCE has a small edge over MSE, as indicated by the results.
The same may also apply to transfer learning of molecule graphs. 
This suggests that a proper criterion is the cornerstone of the success of generative graph self-supervised learning.

Figure \ref{fig:ablation} shows the influence of scaling factor $\gamma$.  We observe that $\gamma > 1$ benefits in most cases, especially in node classification. However, in IMDB-B, it seems that a larger $\gamma$ value harms the performance. In our experiments, we noticed that the training error in node classification is much higher than that in graph classification. This further demonstrates that aligning continuous vectors in unit sphere is more challenging thus scaling $\gamma$ brings improvement.

\vpara{Effect of decoder type.} Previous GAEs typically follow BERT and use an MLP as the decoder. In Table \ref{tab:ablation}, we compare different decoder types, including MLP/GCN/GAT for node classification, and MLP/GIN/GAT for graph classification. The re-mask strategy is not used for MLP decoder.  The type of decoder matters in the training, and using GNN decoder typically achieves better performance in downstream tasks. Compared to MLP, which reconstructs original node feature from latent representation, GNN aims at forcing the masked nodes to extract information about its original feature from neighbors' representations. One reasonable assumption is that GNN avoids the output of encoder being aligned with original features. The learned representations is expected to be discriminative in downstream classification tasks. Therefore, the encoder could learn meaningful latent representations. MLP also works in our framework, and this may attribute to the design of our criterion.

Among different GNNs, GIN performs better in graph-level tasks, and GAT is a more reasonable option for node classification. It is observed that replacing GAT with GCN causes a significant drop, especially in Cora ($\sim$2.9\%) and PubMed ($\sim$2.0\%). We speculate that attention mechanism plays an important role when the latent representations vary in neighbors with re-masked ones.

\begin{table}[htbp]
\caption{Ablation study of decoder type, re-mask and reconstruction criterion on node- and graph-level benchmarks.}
\label{tab:ablation}
\renewcommand\tabcolsep{4pt}

\input{tables/ablation}
\end{table}

\vpara{Effect of mask and re-mask.} \textit{Mask} plays an important role in our method. \model employ two mask strategies special --- masking input feature before encoder, and \textit{re-mask} latent representations before decoder. Table \ref{tab:ablation} studies the designs. We observe a significant drop in performance if without masking input features, indicating that input masking is vital to avoid trivial solution. For the re-mask strategy, the accuracy drops by 0.1\%$\sim$2.9\% if removing the operation. In such case, it behaves like training a $K+1$ layer GNN and adopting the output of the $K$-th layer as the encoding representations. Re-mask is designed for GNN decoder and could be regarded as regularization, making the self-supervisory task more challenging.

\vpara{Effect of mask ratio.} Figure \ref{fig:ablation} shows the influence of mask ratio. Similar to the observation in ~\cite{he2021masked}, in most cases, the reconstruction task is not challenging enough to learn useful features with low mask ratio (0.1), and would be too difficult if the mask ratio is too high (0.9). Different from the behavior in BERT and MAE, the optimal ratio varies in different graphs. In Cora and IMDB-B, increasing the mask ratio degrades the performance after the optimal value ($\sim$0.5), while \model still works with a surprisingly high ratio (0.7$\sim$0.9) in PubMed, Ogbn-arxiv and MUTAG. This motivates us to think about information redundancy in graphs. Large node degree and high homogeneity may lead to heavy information redundancy, in which a missing node feature can be recovered from very few neighboring nodes with little high-level understanding of local structure and features. Masking a very high portion of node features is necessary to learn useful representations. In contrast, lower redundancy means excessively high mask ratio would make it impossible to recover features and degrade the performance. As for how to measure information redundancy, we leave it for future work.

\begin{figure}
    \centering
    \begin{minipage}[t]{0.235\textwidth}
        \includegraphics[width=\textwidth]{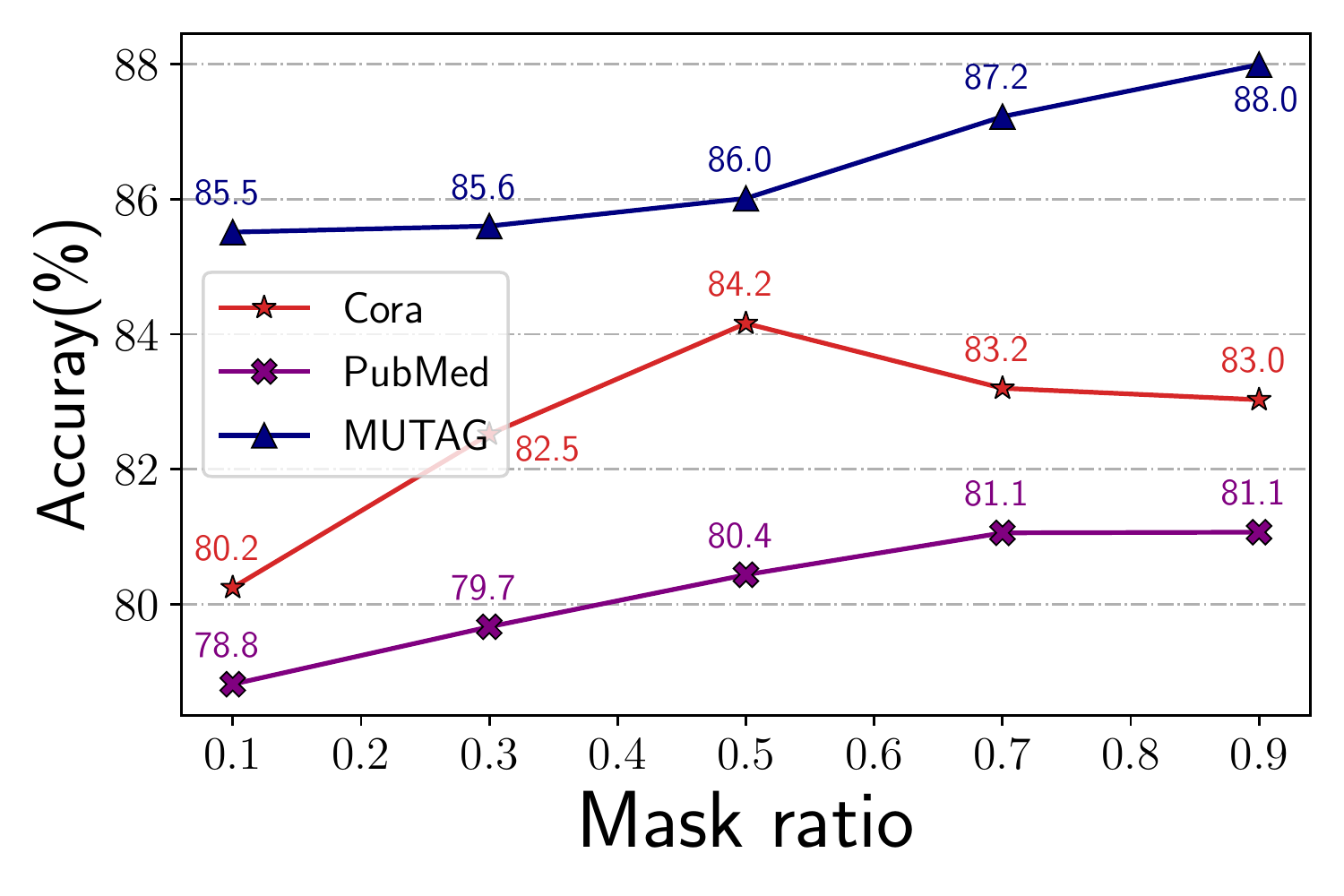}
    \end{minipage}
    \begin{minipage}[t]{0.235\textwidth}
        \includegraphics[width=\textwidth]{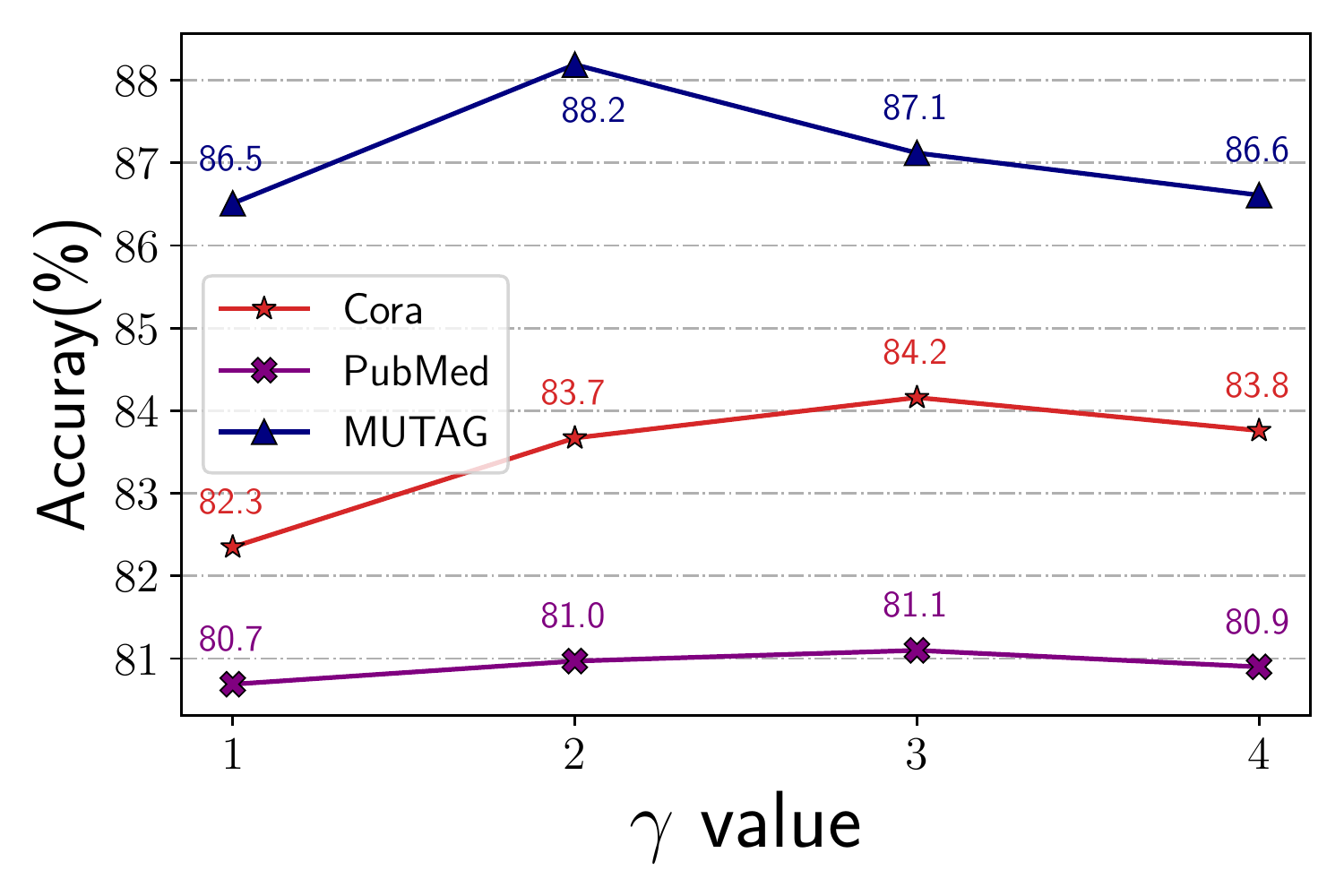}
    \end{minipage}
    \caption{Ablation study of mask ratio and scaling factor $\gamma$.}
    \label{fig:ablation}
\end{figure}


}

%% file: tables/ablation.tex


\begin{tabular}{cccccccc}
\toprule[1.2pt]
\multicolumn{2}{c}{\multirow{2}{*}{Dataset}}
                     & \multicolumn{3}{c}{Node-Level} & \phantom{} & \multicolumn{2}{c}{Graph-Level} \\
\cmidrule{3-5} \cmidrule{7-8}
                    & & Cora        & PubMed      & Arxiv  && MUTAG              & IMDB-B              \\

\midrule
\multirow{4}{*}{\rotatebox[origin=c]{90}{\small{COMP.}}}
& \model        & 84.2  & 81.1  & 71.75  && 88.19         & 75.52         \\
& \small{w/o mask} &  79.7 & 77.9 & 70.97 & & 82.58 & 74.42 \\
& \small{w/o re-mask}   & 82.7  & 80.0  & 71.61  && 86.29         & 74.42          \\
& \small{w/ MSE}                & 79.1  & 73.1  & 67.44  && 86.30         & 74.04          \\
\midrule
\multirow{4}{*}{\rotatebox[origin=c]{90}{\small{Decoder}}}
& MLP                   & 82.2  & 80.4  & 71.54  && 87.16         & 73.94          \\
& GCN                   & 81.3  & 79.1  & 71.59  && 87.78         & 74.54             \\
& GIN                   & 81.8  & 80.2  & 71.41  && 88.19         & 75.52          \\
& GAT                   & 84.2  & 81.1  & 71.75  && 86.27         & 74.04           \\
\bottomrule[1.2pt]
\end{tabular}

%% file: 5.conclusion.tex
\section{Conclusion}%
\label{sec:conclusion}
In this work, we explore generative self-supervised learning in graphs and identify 
the common issues that are faced by graph autoencoders. 
We present \model---a simple masked graph autoencoder---to address them from the  perspective of reconstruction objective, learning, loss function, and model architecture.
In \model, we design the masked feature reconstruction strategy with a scaled cosine error as the reconstruction criterion.  
We conduct extensive experiments on a wide range of node and graph classification benchmarks, and the results demonstrate the  effectiveness and generalizability of \model. 
Our work shows that generative SSL can offer great potential to graph representation learning and pre-training, requiring  
more in-depth explorations for future work.

\hide{

\section{Conclusion}%
\label{sec:conclusion}
In this work, we explore generative learning in graphs and present the challenges of current graph self-supervised learning. Then we propose \model a masked graph autoencoder, to address the problems from the perspective of objective, learning, loss function, and model architecture.
We conduct extensive experiments on a wide range of node and graph classification benchmarks, and the results demonstrate the generalizability and effectiveness of \model. 
Our work shows that generative SSL can offer great potential to graph representation learning and deserve more in-depth exploration in the future.

}

%% file: appendix.tex
\section{Appendix}
\label{app:appendix}

\begin{table}[htbp]
    \centering
    \caption{Comparison with other attributed-masking methods in node classification. ``FT" means finetuning the model in downstream tasks, while ``LP" represents training a linear classifier for classification.}
    \begin{threeparttable}
    \begin{tabular}{c|ccc|c}
        \toprule[1.2pt]
                     & Cora     & Citeseer   & PubMed  & Type   \\
        \midrule
        GAE          & 71.5 &   65.8    & 72.1 & LP \\
        GPT-GNN      &  80.1    & 68.4 &     76.3       & LP \\
        NodeProp     &  81.94   &  71.60     &  79.44   & FT \\
        RASSL-GCN$^{1}$    &  83.80   &  72.95     & *81.23   & FT \\
        GATE         &  83.2    &  71.8      &  80.9    &  LP    \\
        \midrule
        \model       & 84.16    & 73.35      & 81.10    & LP  \\
        \bottomrule[1.2pt]
    \end{tabular}
            \begin{tablenotes}
            \footnotesize
            \item[1] PubMed dataset used is different from that in other baselines.
        \end{tablenotes}
    \end{threeparttable}
    \label{tab:com_attm}
\end{table}

\subsection{Results in the PPI Dataset}
\label{app:ppi}
\begin{figure}[htbp]
    \centering
    \includegraphics[width=0.37\textwidth]{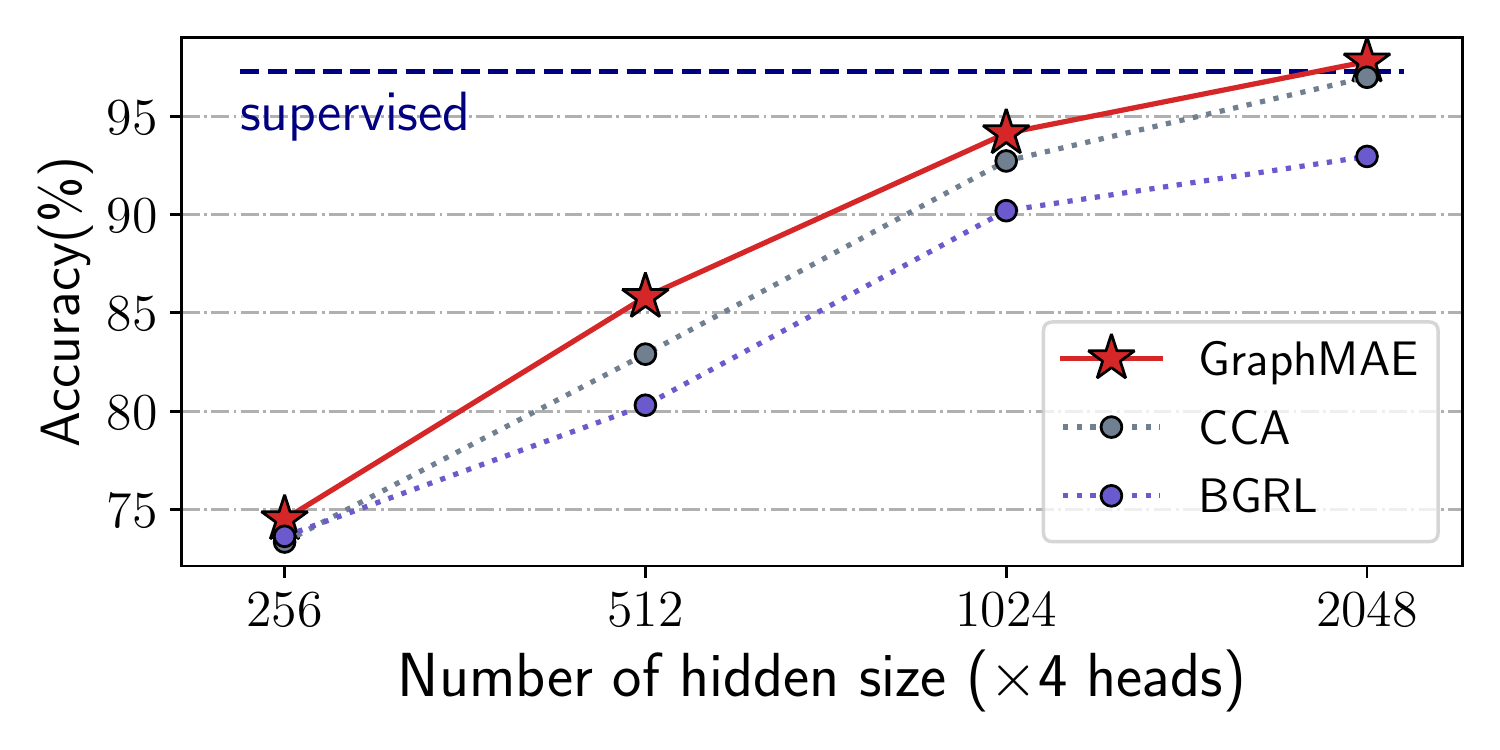}
    \caption{Performance on PPI using GAT with 4 attention heads, compared to other baselines. Self-supervised methods benefit much from larger model size, and \model could outperform supervised model. }
    \label{fig:ppi_hidden}
\end{figure}

Figure \ref{fig:ppi_hidden} shows the results of self-supervised learning methods in PPI dataset. Previous studies often report a large gap between supervised and self-supervised results on PPI. We find that increasing the number of model parameters helps little in supervised setting, but could highly boost the performance in self-supervised setting. As the hidden size reaches 2048$\times$4, \model even outperforms supervised counter-part, although the model would be too much larger. At least, this indicates that the gap could be filled.

The results of BGRL and GRACE in PPI are a bit different from those reported in ~\cite{thakoor2021bootstrapped}. This is because we train the linear classifier until convergence during evaluation, using the authors’ official code. In ~\cite{thakoor2021bootstrapped}, the classifier is trained only for a small fixed number of epochs, and the training does not converge.

\begin{table}[htbp]
\centering
\caption{Experiment results using different encoder backbones in node classification.}
\renewcommand\tabcolsep{3pt}
\begin{tabular}{c|cccc}
\toprule[1.2pt]
          & Cora & Citeseer & Pubmed & Ogbn-arxiv \\
\midrule
BGRL (GCN)    &   82.7$\pm$0.6    &  71.1$\pm$0.7        & 79.4$\pm$0.6       &  71.64$\pm$0.12      \\
BGRL (GAT)    &   82.8$\pm$0.5    &   71.1$\pm$0.8       &  79.6$\pm$0.5      &   70.07$\pm$0.02 \\
CCA-SSG (GCN) &   84.0$\pm$0.4      &  73.1$\pm$0.3         &  81.0$\pm$0.4        &  70.81$\pm$0.13  \\
CCA-SSG (GAT) &   83.8$\pm$0.5    &  72.6$\pm$0.7        &   79.9$\pm$1.1     &  71.24$\pm$0.20  \\
\model (GCN)  &   82.9$\pm$0.6             &  72.5$\pm$0.5                 &  81.0$\pm$0.5               &  \bf 71.87$\pm$0.21\\   
\model (GAT)  &   \bf 84.2$\pm$0.4        &  \bf 73.4$\pm$0.4       & \bf 81.1$\pm$0.4      &  71.75$\pm$0.17 \\
\bottomrule[1.2pt]
\end{tabular}
\vspace{-2mm}
\label{tab:backbone}
\end{table}

\subsection{Ablation on the Encoder Architecture}
In node classification, \model uses GAT as the encoder. To have a fair comparison and investigate the influence of different GNN backbones, we compare the best baselines, BGRL and CCA-SSG, in node classification datasets using GCN and GAT as the encoder. The results are shown in Table \ref{tab:backbone}. We observe that \model still outperforms the baselines with the same GAT backbone. 
In addition, the results manifest that attention mechanism would not always benefit graph self-supervised learning, as GCN is inferior to GAT, for instance, in (Cora, CCA-SSG) and (Ogbn-arxiv, \model). Under the training setting of \model, GAT could be a better option in most cases.

\subsection{Implementation Details}
\subsubsection{Environment}
Most experiments are conducted on Linux servers equipped with an Intel(R) Xeon(R) CPU E5-2680 v4 @ 2.40GHz, 256GB RAM and NVIDIA 2080Ti GPUs. Experiments of Ogbn-arxiv and Reddit for node classificatin are conducted on Intel(R) Xeon(R) Gold 6240 CPU @ 2.60GHz and NVIDIA 3090 GPUs, as they require large memory. Models of node and graph classification are implemented in PyTorch version 1.9.0, DGL version 0.7.2 (\textit{\url{https://www.dgl.ai/}})  with CUDA version 10.2, scikit-learn version 0.24.1 and Python 3.7. 
For molecular property prediction, we implement our model based on the code in \textit{\url{https://github.com/snap-stanford/pretrain-gnns}} with Pytorch Geometric 2.0.4 (\textit{\url{https://www.pyg.org/}}).

\begin{figure}[htbp]
    \centering
    \begin{minipage}{0.23\textwidth}
        \includegraphics[width=\textwidth]{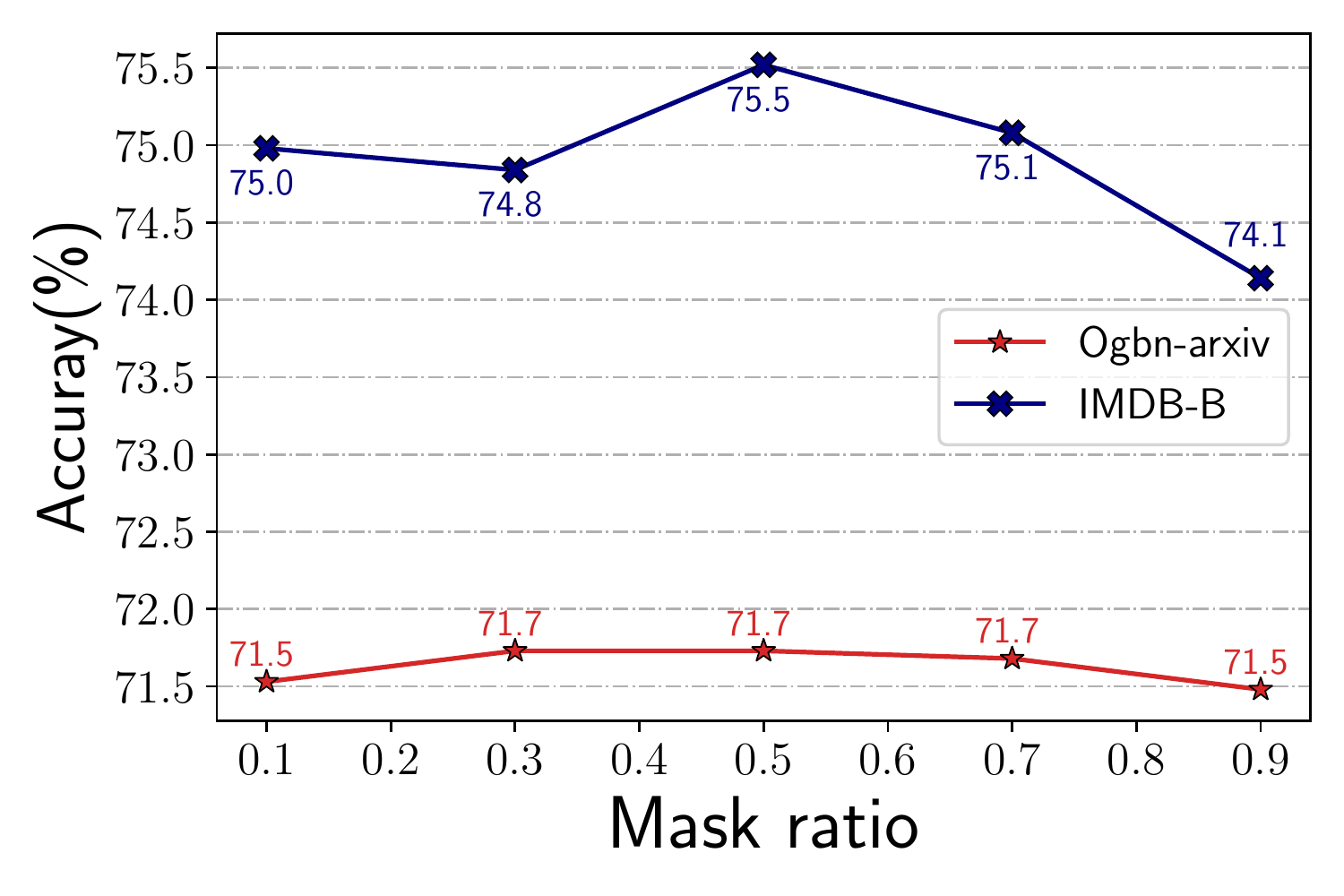}
    \end{minipage}
    \begin{minipage}{0.23\textwidth}
        \includegraphics[width=\textwidth]{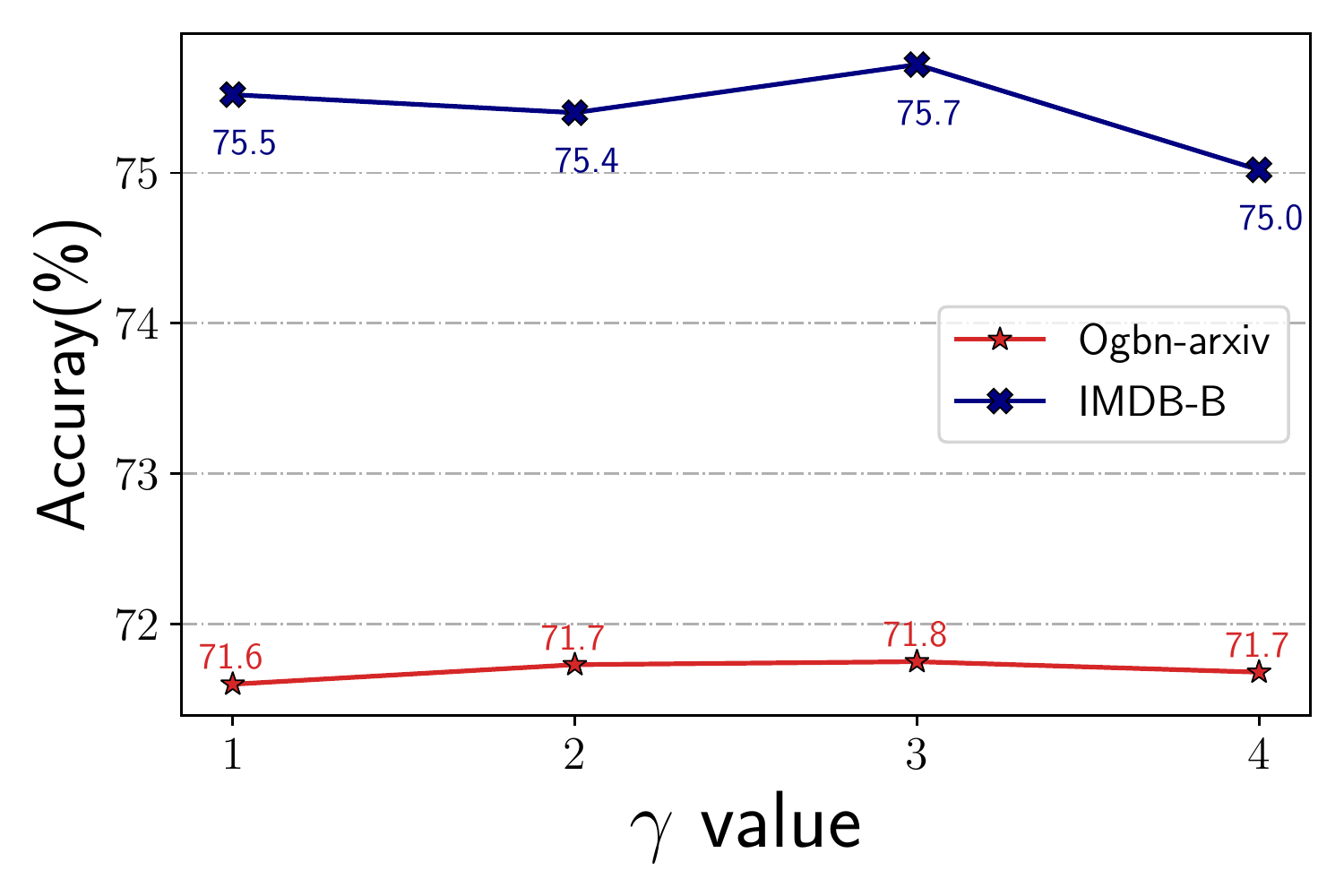}
    \end{minipage}
    \caption{Ablation study of mask ratio and scaling factor $\gamma$ in Ogbn-arxiv and IMDB-B.}
    \label{fig:abl_append}
\end{figure}

\hide{
\subsubsection{Model Configuration}
For node classification, we train the model using Adam Optimizer with $\beta_1 = 0.9, \beta_2 = 0.999, \epsilon = 1 \times 10^{-8}$. The initial learning rate is set to 0.001, using cosine learning rate decay without warmup. We use PReLU as the non-linear activation. More details about hyper-parameters and datasets are in Table \ref{tab:hp_node} and \url{https://github.com/THUDM/GraphMAE}.

For graph classification, we set the initial learning rate to 0.00015 with cosine learning rate decay for most cases. For the evaluation,  the parameter C of SVM is searched in the sets $\{10^{-3},...,10\}$. The hyper-parameters and statistics of datasets are in Table ~\ref{tab:hp_graph}.

For transfer learning of molecule property prediction, we adopt a single-layer GIN as decoder, the mask rate is set to 0.25, and we pretrain the whole model for 100 epochs.
For the finetuning, an Adam optimizer (learning rate: 0.001, batch size: 32) is employed to train the model for 100 epochs. We utilize a learning rate scheduler with fix step size, which multiplies the learning rate by 0.3 every 30 epochs for BACE only. Table ~\ref{tab:stat_mol} shows the statistics of datasets.

\subsection{Baselines}
For node classification, the results of supervised baselines of GCN in Reddit, and GAT in Ogbn-arxiv and Reddit are from CogDL(\textit{\url{https://cogdl.ai/}}) if not reported before. For unsupervised baselines, GRACE~\cite{zhu2020deep}, BGRL~\cite{thakoor2021bootstrapped}, CCA-SSG~\cite{zhang2021canonical} are state-of-the-art contrastive learning methods in graph. 
GRACE and BGRL did not report the results in Cora, Citeseer, and PubMed of the public split. To have a fair comparison, we download the public source code and use the same GNN backbone as \model. We conduct hyper-parameter search for them and select the best results on the validation set.
The results of CCA-SSG are the output of the official code after the bugs in the code are fixed. For MVGRL~\cite{hassani2020contrastive}, we adopt DGL's reproducing results. We implement GPT-GNN~\cite{hu2020gpt} based on the official code for homogeneous networks, as the code is for heterogeneous networks, and report the results in Cora, Citeseer, and PubMed. 

For graph classification and transfer learning of molecular property prediction, we adopt the results reported in previous papers if available. For the results of GraphCL~\cite{you2020graph} and JOAO~\cite{you2021graph} in IMDB-MULTI, we download the authors' official codes and keep hyper-parameters the same to get the output. }

\begin{table*}[htbp]
\centering
\label{tab:graph_sta_hyper}
\caption{Statistics and hyper-parameters for node classification datasets. ``s'' represents multi-class classification, and ``m'' means multi-label classification. 
}
\renewcommand\tabcolsep{6.5pt}
\renewcommand\arraystretch{1.05}
\begin{tabular}{c|c|cccccc}
\toprule[1.2pt]
& Dataset               & Cora & Citeseer & PubMed & Ogbn-arxiv & PPI       & Reddit    \\
\midrule
\multirow{3}{*}{Statistics}
&         \# nodes   & 2,708 & 3,327 & 19,717 & 169,343      & 56,944  & 232,965      \\
&         \# edges   & 5,429 & 4,732 & 44,338 & 1,166,243    & 818,736 & 11,606,919   \\
&         \# classes & 7(s)  & 6(s)  & 3(s)   & 40(s)        & 121(m)  & 41(s)         \\
\midrule
\multirow{6}{*}{Hyper-parameters} 
& scaling factor $\gamma$ & 3    & 1        & 3      & 3          & 3         & 3         \\
& masking rate            & 0.5  & 0.5      & 0.75   & 0.5        & 0.5       & 0.75      \\
& replacing rate          & 0.05 & 0.10     & 0      & 0          & 0         & 0.15      \\
& hidden\_size            & 512  & 512      & 1024   & 1024       & 1024      & 512       \\
& weight\_decay           & 2e-4 & 2e-5     & 1e-5   & 0          & 0         & 2e-4      \\
& max\_epoch              & 1500 & 300      & 1000   & 1000       & 1000      & 500       \\
\bottomrule[1.2pt]
\end{tabular}
\label{tab:hp_node}
\end{table*}

\begin{table*}[htbp]
\centering
\label{tab:node_sta_hyper}
\caption{Statistics and hyper-parameters for graph classification datasets. 
}
\renewcommand\arraystretch{1.05}
\begin{tabular}{c|c|ccccccc}
        \toprule[1.2pt]
         &  Dataset             & IMDB-B    & IMDB-M    & PROTEINS  & COLLAB    & MUTAG     & REDDIT-B  & NCI1      \\
        \midrule
\multirow{3}{*}{Statistics}             

        & \# graphs      & 1,000 & 1,500 & 1,113 & 5,000 & 188  & 2,000 & 4,110 \\     
        & \# classes     &  2    & 3     & 2     & 3     & 2    & 2     & 2    \\            
        & Avg. \# nodes  & 19.8  & 13.0  & 39.1  & 74.5  & 17.9 & 429.7 & 29.8  \\
        \midrule
\multirow{7}{*}{\begin{tabular}[c]{@{}c@{}}Hyper-\\ parameters\end{tabular}} 
         & Scaling factor $\gamma$ & 1         & 1         & 1         & 1         & 2         & 1         & 2         \\
         & masking rate            & 0.5       & 0.5       & 0.5       & 0.75      & 0.75      & 0.75      & 0.25      \\
         & replacing rate          & 0.0       & 0.0       & 0.0       & 0.0       & 0.1       & 0.1       & 0.1       \\
         & hidden\_size            & 512       & 512       & 512       & 512       & 32        & 512       & 512       \\
         & weight\_decay           & 0         & 0         & 0         & 0         & 0.0       & 0.0       & 0         \\
         & max\_epoch              & 60        & 50        & 100       & 20        & 20        & 100       & 300       \\
         & batch\_size             & 32        & 32        & 32        & 32        & 64        & 8         & 16         \\
         & Pooling                 & mean      & mean      & max       & max       & sum       & max       & sum        \\
        \bottomrule[1.2pt]
\end{tabular}
\label{tab:hp_graph}
\end{table*}

\begin{table*}[htbp]
    \centering
    \renewcommand\arraystretch{1.05}
    \caption{Statistics of datasets for molecular property prediction. ``ZINC'' is used for pre-training.}
    \begin{tabular}{@{}c|ccccccccc}  
        \toprule[1.2pt]
                      & ZINC & BBBP & Tox21 & ToxCast & SIDER & ClinTox & MUV & HIV & BACE \\
        \midrule
        \# graphs      & 2,000,000 & 2,039 & 7,831 & 8,576 & 1,427 & 1,477 & 93,087 & 41,127 & 1,513\\     
        \# binary prediction tasks     & - & 1 & 12 & 617 & 27 & 2 & 17 & 1 & 1   \\            
        Avg. \# nodes  & 26.6 & 24.1 & 18.6 & 18.8 &  33.6 & 26.2 & 24.2 & 24.5 & 34.1 \\ 
        \bottomrule[1.2pt]
    \end{tabular}
    \label{tab:stat_mol}
\end{table*}

\subsubsection{Model Configuration}
For node classification, we train the model using Adam Optimizer with $\beta_1 = 0.9, \beta_2 = 0.999, \epsilon = 1 \times 10^{-8}$. The initial learning rate is set to 0.001, using cosine learning rate decay without warmup. We use PReLU as the non-linear activation. More details about hyper-parameters and datasets are in Table \ref{tab:hp_node} and \url{https://github.com/THUDM/GraphMAE}.

For graph classification, we set the initial learning rate to 0.00015 with cosine learning rate decay for most cases. For the evaluation,  the parameter C of SVM is searched in the sets $\{10^{-3},...,10\}$. The hyper-parameters and statistics of datasets are in Table ~\ref{tab:hp_graph}.

For transfer learning of molecule property prediction, we adopt a single-layer GIN as decoder, the mask rate is set to 0.25, and we pretrain the whole model for 100 epochs.
For the finetuning, an Adam optimizer (learning rate: 0.001, batch size: 32) is employed to train the model for 100 epochs. We utilize a learning rate scheduler with fix step size, which multiplies the learning rate by 0.3 every 30 epochs for BACE only. Table ~\ref{tab:stat_mol} shows the statistics of datasets.

\subsection{Baselines}
For node classification, the results of supervised baselines of GCN in Reddit, and GAT in Ogbn-arxiv and Reddit are from CogDL(\textit{\url{https://cogdl.ai/}}) if not reported before. For unsupervised baselines, GRACE~\cite{zhu2020deep}, BGRL~\cite{thakoor2021bootstrapped}, CCA-SSG~\cite{zhang2021canonical} are state-of-the-art contrastive learning methods in graph. 
GRACE and BGRL did not report the results in Cora, Citeseer, and PubMed of the public split. To have a fair comparison, we download the public source code and use the same GNN backbone as \model. We conduct hyper-parameter search for them and select the best results on the validation set.
The results of CCA-SSG are the output of the official code after the bugs in the code are fixed. For MVGRL~\cite{hassani2020contrastive}, we adopt DGL's reproducing results. We implement GPT-GNN~\cite{hu2020gpt} based on the official code for homogeneous networks, as the code is for heterogeneous networks, and report the results in Cora, Citeseer, and PubMed. 

For graph classification and transfer learning of molecular property prediction, we adopt the results reported in previous papers if available. For the results of GraphCL~\cite{you2020graph} and JOAO~\cite{you2021graph} in IMDB-MULTI, we download the authors' official codes and keep hyper-parameters the same to get the output. 